\documentclass[11pt]{article}

\usepackage[T1]{fontenc}
\usepackage[utf8]{inputenc}
\usepackage{lmodern}
\usepackage[margin=1in]{geometry}
\usepackage{amsmath}
\usepackage{amssymb}
\usepackage{booktabs}
\usepackage{graphicx}
\usepackage{xcolor}
\usepackage{microtype}
\usepackage[numbers,sort&compress]{natbib}
\usepackage[hidelinks]{hyperref}

\title{MiniVer-V: Identifying Minimal Sufficient Evidence\\for Short Video Verification}

\author{
  Leran Chen \\
  The University of Sydney \\
  {\small\texttt{lche0711@uni.sydney.edu.au}}
  \and
  Lingnan Kong \\
  The University of Sydney \\
  {\small\texttt{lkon0802@uni.sydney.edu.au}}
  \and
  Zile Cai \\
  The University of Sydney \\
  {\small\texttt{zcai0293@uni.sydney.edu.au}}
}

\date{}

\begin{document}

\maketitle

\begin{abstract}
A core challenge in short-video fact-checking is identifying which evidence is sufficient to support a verification conclusion. Existing approaches either give the verifier all available evidence, introducing noise, or select evidence by topical relevance, which conflates relatedness with sufficiency. We identify evidential sufficiency as the selection criterion: whether a subset of evidence is adequate to support a confident verdict without redundancy. We introduce \textbf{MiniVer-V}, a benchmark of 195 short videos with three-way verdict annotations (supported, refuted, insufficient) and 5,510 multimodal evidence units spanning visual keyframes, speech transcripts, and web-retrieved external sources. We propose a two-layer verification framework that separates claim--video consistency, assessed from internal evidence, from factual verdict determination, which additionally requires external corroboration. On top of it, a sufficiency-driven greedy search assembles evidence until a sufficiency threshold is met and outputs \textit{insufficient} when the candidate pool is exhausted, rather than forcing a verdict. With Claude Sonnet 4, the method reaches a Macro-F1 of 0.510 using 4.5 evidence units on average (16\% of the full evidence set), statistically indistinguishable from the full-evidence baseline (0.518 with 27.7 units), while significantly improving recognition of insufficient cases over the same search without abstention. The efficiency result replicates with GPT-5.5 and holds only partially with an open-weight Qwen2.5-72B verifier. Ablations show that external evidence is indispensable for factual determination, while internal video evidence grounds the verdict in claim--video consistency. These findings suggest that evidence-efficient verification is achievable, and that explicit abstention is needed when evidence is genuinely inadequate.
\end{abstract}


\section{Introduction}
\label{sec:introduction}

Short-video platforms have become a major channel for news consumption and a breeding ground for misinformation~\cite{qi2023fakesv,bu2024fakingrecipe}. Unlike textual claims or static image--text posts~\cite{shao2023dgm4}, a short video weaves together visual scenes, spoken narration, overlaid text, and contextual metadata into a tightly coupled multimodal package that is persuasive, rapidly consumable, and easy to redistribute~\cite{bu2024fakingrecipe}. Verifying the factual claims embedded in such videos requires reasoning over heterogeneous evidence---yet not all evidence contributes equally. Consider a video that accurately depicts a real flood: the footage is genuine, the narration matches the visuals, and the claim appears credible. Yet the claim misattributes the footage to a different city and a different year. Visual keyframes confirm the event occurred but cannot establish \emph{where} or \emph{when}; only an external news report from the claimed date and location can confirm or deny the attribution. The critical question is not which evidence is \emph{relevant}, but which evidence is \emph{sufficient}.

Existing fact-checking and multimodal verification systems typically either aggregate all available evidence or rank evidence by relevance and select a fixed-budget subset~\cite{thorne2018fever,yao2023end,qi2023fakesv}. Both paradigms implicitly assume that more---or more relevant---evidence leads to more accurate verdicts. We identify three limitations of this assumption. First, relevance and sufficiency are fundamentally different properties: a set of highly relevant units may collectively fail to establish the key factual link needed for verification, while a moderately relevant unit may provide the missing piece. Second, evidence beyond what is needed introduces redundancy and noise that can impair reasoning quality~\cite{shi2023distracted,zheng2024evidence}. Third, and most critically, systems without an explicit abstention mechanism tend to issue verdicts even when the evidence is insufficient~\cite{atanasova2022fact}---a failure mode especially dangerous in misinformation detection, where a confident but wrong verdict can be more harmful than abstaining. Despite these risks, to our knowledge no existing multimodal video verification benchmark explicitly evaluates whether a model's chosen evidence is truly sufficient and free of redundancy.

In this paper, we formalize \emph{minimal sufficient evidence set identification} as a new task for short video verification. Given a claim and a pool of multimodal evidence units derived from the video and from external sources, the model must select a compact cross-modal evidence subset sufficient to support a confident verdict---and explicitly abstain when no such subset exists. This formulation differs from conventional evidence ranking in two key respects: sufficiency is a first-class objective rather than an implicit by-product of relevance, and evidence selection is jointly constrained by sufficiency and compactness---the selector prefers subsets whose elements are not redundant with one another.

To study this task, we construct \textbf{MiniVer-V}, a benchmark of 195 short videos with three-way verdict annotations (supported, refuted, insufficient) and 5,510 multimodal evidence units spanning visual keyframe descriptions (V), speech transcripts (A), and web-retrieved external sources (E). We propose a two-layer verification framework: the first layer assesses claim--video consistency using internal evidence (V and A) alone; the second incorporates external evidence to reach a factual verdict. We evaluate five evidence selection strategies of increasing sophistication---from full-evidence and fixed-budget top-$k$ baselines, through model-driven chain-of-thought selection, to our proposed method: a sufficiency-driven greedy search with explicit abstention (B5) that dynamically determines how much evidence is needed per sample and outputs \textit{insufficient} when the candidate pool is exhausted without reaching the sufficiency threshold.

Experiments with Claude Sonnet 4 show that B5 achieves a Macro-F1 of 0.510 using 4.5 evidence units on average (16\% of the full evidence set), recovering 98.5\% of full-evidence performance. Compared with the greedy variant without abstention, B5 nearly doubles recall on insufficient cases. The efficiency result replicates with GPT-5.5 and holds only partially with the open-weight Qwen2.5-72B. Ablations show that external evidence is indispensable for factual determination, while internal evidence remains necessary for claim--video grounding. This work makes three contributions:
\begin{itemize}
    \item We introduce \emph{minimal sufficient evidence set identification} as a task for short video verification, highlighting the underexplored gap between evidence relevance and evidence sufficiency.
    \item We construct MiniVer-V, a benchmark of 195 short videos with 5,510 multimodal evidence units that enables controlled experiments on evidence selection strategies for video verification.
    \item We show that sufficiency-aware selection with explicit abstention recovers 98.5\% of full-evidence performance using only 16\% of the evidence while improving recognition of genuinely insufficient cases.
\end{itemize}

\section{Related Work}
\label{sec:related_work}

\subsection{Multimodal Misinformation Detection}
\label{sec:rw_misinformation}

Recent work has advanced short-video misinformation detection through cross-modal benchmarks and video--text verification datasets~\cite{qi2023fakesv,bu2024fakingrecipe,zhang2025fakevv,zeng2026wildfakebench}. In the broader fact-checking literature, FEVER, MOCHEG, and AVeriTeC address text-based, multimodal, and web-grounded settings respectively~\cite{thorne2018fever,yao2023end,schlichtkrull2023averitec}, while DEFAME dynamically plans which multimodal tools to invoke and how much evidence to retrieve~\cite{braun2025defame}.

These works evaluate final verdict accuracy, and most of them provide evidence as a fixed input or retrieve it as a single package. To our knowledge, none treats evidence selection---how much evidence is enough and which subsets suffice---as a primary evaluation objective. MiniVer-V addresses this gap by providing unit-level evidence decomposition and evaluating selection strategies under controlled conditions.

\subsection{Evidence Retrieval and Selection}
\label{sec:rw_evidence}

Automated fact-checking is commonly framed as a pipeline of claim detection, evidence retrieval, and verdict prediction, with the reliability of retrieved evidence identified as a central open challenge~\cite{guo2022survey}. Structured reasoning over retrieved evidence~\cite{liu2020kgat}, claim decomposition into sub-tasks~\cite{pan2023fact}, and improved retrieval strategies~\cite{zheng2024evidence} have all yielded substantial gains. However, these approaches focus on evidence \emph{relevance}---retrieving evidence topically related to the claim. Relevance and sufficiency are distinct: a set of highly relevant units may still fail to establish the critical factual link needed for verification. Atanasova et~al.~\cite{atanasova2022fact} formalize fact-checking with insufficient evidence and show that standard models often fail to recognize when evidence is insufficient. Our work extends this insight to the multimodal video setting by treating sufficiency as a first-class selection criterion.

\subsection{Selective Prediction and Abstention}
\label{sec:rw_abstention}

Selective prediction---abstaining when confidence is low or evidence is inadequate---has a long history in machine learning, with a recent survey framing LLM abstention from query, model, and human-value perspectives~\cite{wen2025know}. In fact verification, the need for abstention arises naturally: FEVER and AVeriTeC both include ``not enough evidence'' verdict categories. Yet most verification systems lack an explicit mechanism to predict these labels reliably, defaulting to a forced choice regardless of evidence quality. Our method addresses this by incorporating a sufficiency threshold: when the greedy search exhausts the candidate pool without reaching sufficiency, the system outputs \textit{insufficient} rather than forcing a potentially unreliable verdict, treating abstention as a procedural outcome of evidence insufficiency rather than a classification label to be learned.

\section{MiniVer-V Benchmark}
\label{sec:benchmark}
\subsection{Task Setting and Benchmark Motivation}
\label{sec:task_setting}

\textbf{Task Definition.}
Given a short video paired with a textual claim about its content, the task is to determine a factual verdict---\textit{supported}, \textit{refuted}, or \textit{insufficient}---by selecting and reasoning over a subset of multimodal evidence units derived from the video itself and from external sources.

Unlike binary fact-checking formulations that assign claims a true-or-false label, we adopt a three-way verdict scheme. The \textit{insufficient} category captures an epistemic state common in real-world verification: the available evidence may be genuinely inadequate to determine whether a claim is true or false. In our benchmark, 30.3\% of samples (59 out of 195) carry this label, showing that evidence insufficiency is a substantial and practically relevant case rather than a marginal exception.

\textbf{Motivation.}
MiniVer-V is designed for controlled study of evidence sufficiency in short-video verification. By decomposing each sample into modality-labeled evidence units, it enables comparison of evidence selection strategies under a shared verification setting.

\subsection{Data Collection}
\label{sec:data_collection}

We curate 195 short videos through two complementary collection strategies: (i) tracing videos referenced in published fact-checking reports back to their original social media posts, and (ii) independently browsing social media platforms to identify videos whose factual claims warrant verification. Candidates are filtered to remove duplicates, videos shorter than 10 seconds, and clips with severely degraded visual quality. We further exclude videos whose claims are purely subjective or for which no verifiable evidence---either internal to the video or externally retrievable---could reasonably be obtained. All claims are written in Chinese. The videos are predominantly Chinese-language; 40 of 195 (20.5\%) contain mainly English speech. Videos come primarily from Douyin (112; 57.4\%) and Xiaohongshu (64; 32.8\%), with the remainder from X, Bilibili, Weibo, and news outlets (Appendix~\ref{app:platform}). Topics span international politics, public health, science, consumer safety, and natural disasters.

For each video, an annotator writes a textual claim capturing its main factual assertion and assigns a verdict based on the video content and web-retrieved external evidence, using the original caption only as auxiliary context. A second team member reviews a random subset of samples to assess annotation consistency, yielding Cohen's $\kappa$ = 0.82.

Each sample is assigned one of three verdict labels: \textit{supported} (the claim is substantiated by the available evidence), \textit{refuted} (the claim is contradicted by the available evidence), or \textit{insufficient} (the evidence does not permit a confident determination). The distribution across the 195 samples is shown in Table~\ref{tab:verdict_dist}.

\begin{table}[htbp]
\centering
\caption{Verdict distribution in MiniVer-V.}
\label{tab:verdict_dist}
\begin{tabular}{lcc}
\toprule
Verdict & Count & Proportion \\
\midrule
Supported & 97 & 49.7\% \\
Refuted & 39 & 20.0\% \\
Insufficient & 59 & 30.3\% \\
\midrule
\textbf{Total} & \textbf{195} & \textbf{100\%} \\
\bottomrule
\end{tabular}
\end{table}

The label distribution is imbalanced, with nearly half of the claims labeled as supported and only one-fifth as refuted. We retain the observed distribution rather than artificially balancing it, as it better reflects real-world verification conditions.

\subsection{Multimodal Evidence Extraction}
\label{sec:evidence_extraction}

Each video is processed into discrete, modality-labeled evidence units, all represented in text form so that heterogeneous evidence can be handled under a unified interface. As a result, visual and audio signals are consumed through model-generated descriptions and transcripts rather than raw inputs. The three evidence modalities and the treatment of metadata are described below.

\textbf{Visual keyframes (V).} Keyframes are extracted by uniform temporal sampling at 5-second intervals, yielding up to 8 frames per video. Each sampled frame is processed by a vision-language model (Claude Sonnet 4), which generates a structured textual description including the scene, salient persons and objects, transcription of on-screen text, and visual cues relevant to fact verification. The resulting V-type unit is this textual description rather than the raw image. The description prompt asks for at most 150 characters to keep visual evidence concise and comparable across samples. This process produces 1,207 units across all samples (mean: 6.2 per sample).

\textbf{ASR transcripts (A).} Spoken content is transcribed using an automatic speech recognition model (Whisper base) with automatic language detection. We use the timestamped transcript segments returned by the model as the A-type evidence units. This yields 2,847 segments in total (mean: 14.6 per sample), making ASR the largest single modality by volume, as spoken content varies substantially in length across videos.

\textbf{External web evidence (E).} External evidence is retrieved via web search. For each claim, two rounds of queries are issued: (1) the original claim text, and (2) the claim augmented with language-matched fact-checking keywords (e.g., ``fact check'' or ``debunk'' in English, and corresponding terms in Chinese). Each round returns up to 5 results, which are then deduplicated by URL. Each retrieved result is represented as a structured E-type unit containing a title, content snippet, and source URL. This yields 1,456 units in total (mean: 7.5 per sample).

\textbf{Metadata exclusion.} Each video also carries a metadata unit containing the video title and platform-level contextual information. However, because the textual claim for each sample is written by the annotator after watching the video (Section~\ref{sec:data_collection}), the video title frequently overlaps with or closely paraphrases the claim itself. Including such metadata as evidence would introduce circular reasoning---the verifier would effectively compare the claim against a near-duplicate of itself. We therefore exclude metadata units from the verification evidence pool. All subsequent statistics, evidence counts, and experiments refer to the three-modality pool (V, A, E) after metadata exclusion.

The complete evidence composition is summarized in Table~\ref{tab:evidence_composition}.

\begin{table}[htbp]
\centering
\caption{Evidence unit composition in MiniVer-V (metadata excluded).}
\label{tab:evidence_composition}
\begin{tabular}{llccr}
\toprule
Modality & Code & Total Units & Avg / Sample & Proportion \\
\midrule
Visual keyframes & V & 1,207 & 6.2 & 21.9\% \\
ASR transcripts & A & 2,847 & 14.6 & 51.7\% \\
External search & E & 1,456 & 7.5 & 26.4\% \\
\midrule
\textbf{Total} & & \textbf{5,510} & \textbf{28.3} & \textbf{100\%} \\
\bottomrule
\end{tabular}
\end{table}

The number of evidence units per sample ranges from 9 to 91, with a median of 26. This unit-level decomposition enables fine-grained control over evidence selection, allowing systematic comparison of strategies that vary in both the amount and composition of evidence presented to the verifier.

\subsection{Dataset Statistics and Comparison}
\label{sec:dataset_stats}

Evidence counts vary substantially across samples (9--91 units; mean 28.3, median 26), with the right-skew driven mainly by ASR length. Internal evidence (V+A) accounts for 73.6\% of all units, making modality imbalance a practical challenge for evidence selection.

Table~\ref{tab:benchmark_comparison} compares MiniVer-V with representative fact-checking benchmarks across key design dimensions.

\begin{table}[htbp]
\centering
\small
\setlength{\tabcolsep}{4pt}
\caption{Comparison of MiniVer-V with representative fact-checking benchmarks. Selection focus: whether evidence selection is the primary object of evaluation. $^{\ast}$FakeSV: fake and real videos used for binary detection (5,538 videos including debunking videos).}
\label{tab:benchmark_comparison}
\begin{tabular}{lllcllc}
\toprule
Benchmark & Media & Samples & Lang & Labels & Evidence (granularity) & Selection focus \\
\midrule
FEVER & Text & $\sim$185K & EN & 3-way & Text sentences & No \\
AVeriTeC & Text & $\sim$4.6K & EN & 4-way & QA pairs + URLs & No \\
MOCHEG & Multimodal web & $\sim$15.6K & EN & 3-way & Paragraphs + images & No \\
FakeSV & Short video & $\sim$3.7K$^{\ast}$ & ZH & Binary & V + A + M (per sample) & No \\
\textbf{MiniVer-V} & \textbf{Short video} & \textbf{195} & \textbf{ZH/EN} & \textbf{3-way} & \textbf{V + A + E (per unit)} & \textbf{Yes} \\
\bottomrule
\end{tabular}
\end{table}

MiniVer-V combines design elements that appear separately in prior benchmarks---three-way labels, short-video content, and web evidence---but differs in making unit-level evidence selection the primary object of study. Although smaller in sample count than existing benchmarks, MiniVer-V is intended as an evaluation benchmark for controlled evidence-selection experiments rather than a large-scale training resource~\cite{roberts2024scifibench}. Here, the key experimental variation lies in the selectable evidence units and the subsets induced per sample.

\subsection{Evidence Quality Control and Annotation}
\label{sec:evidence_qc}

A random 20\% of samples undergo manual inspection to verify extraction quality; units that fail are re-extracted or corrected. Each evidence unit is additionally assigned an automatically generated relation label (\textit{support}, \textit{refute}, or \textit{irrelevant}) with respect to the claim, produced by the backbone model with a 20\% human audit. These relation labels are not used as input to any verification strategy; they serve as metadata for post-hoc analysis.

\subsection{Human Annotation of Reference Evidence Sets}
\label{sec:human_annotation}

To validate automated evidence selection, three graduate-student annotators independently select compact evidence subsets they consider sufficient for the ground-truth verdict. Annotators are shown the claim, ground-truth verdict, and all evidence units grouped by modality. We obtain annotations for all 195 samples. For evaluation, we construct a single reference set for each sample by retaining evidence units selected by at least two annotators. Quantitative comparison with B5 is reported in Section~\ref{sec:human_validation}.

\section{Method}
\subsection{Overview}
\label{sec:method_overview}

\begin{figure*}[t]
\centering
\includegraphics[width=\textwidth]{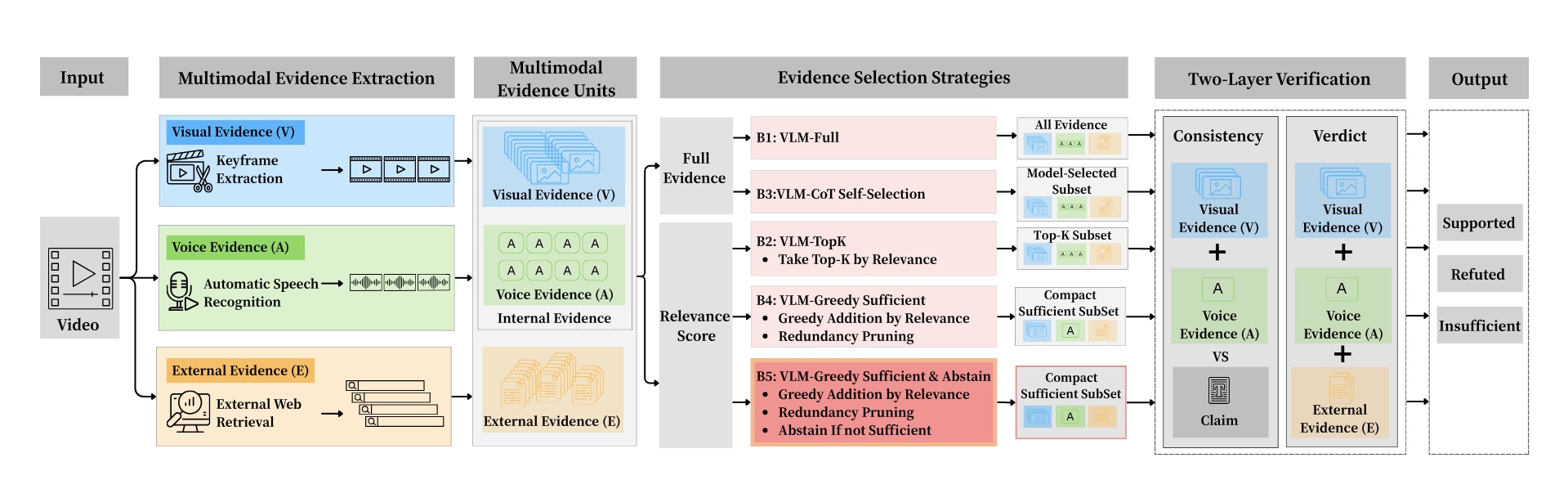}
\caption{Overview of the MiniVer-V verification framework. Given a claim and a multimodal evidence pool (visual keyframes, speech transcripts, and web-retrieved external sources), five evidence selection strategies of increasing sophistication are evaluated. B5, our proposed method, performs sufficiency-driven greedy search with explicit abstention, feeding the selected compact evidence subset into a two-layer verifier that disentangles claim--video consistency (Layer~1) from factual verdict determination (Layer~2).}
\label{fig:framework}
\end{figure*}

Given a claim $c$ and its associated set of evidence units $\mathcal{E} = \{e_1, e_2, \ldots, e_n\}$ extracted in Section~\ref{sec:evidence_extraction}, the verification task proceeds in three stages: (1) relevance scoring and candidate pool construction, (2) evidence selection, and (3) two-layer verdict prediction. The output is a selected subset $\mathcal{S} \subseteq \mathcal{E}$ together with a factual verdict. Figure~\ref{fig:framework} illustrates the overall architecture.

We instantiate the evidence selection stage with five strategies that form a progression of increasing sophistication. B1 (Full Evidence) uses all available units without selection, serving as the full-evidence reference. B2 (Top-K) selects a fixed number of highest-scoring units, testing whether relevance ranking alone yields a sufficient subset. B3 (CoT Self-Selection) asks the model to select a subset via chain-of-thought reasoning. B4 (Greedy-Sufficient) introduces a sufficiency-aware search loop that dynamically determines how many units are needed per sample, followed by redundancy pruning. B5 (Greedy + Abstention), our proposed method, extends B4 by outputting \textit{insufficient} when the candidate pool is exhausted without reaching sufficiency, rather than forcing a verdict. This progression---from no selection, through fixed-budget and model-driven selection, to sufficiency-driven dynamic selection with explicit abstention---is designed to isolate the contribution of each design choice.

All five strategies share the same backbone model, prompt family, and two-layer verifier; they differ only in how evidence is selected, with B5 additionally introducing an abstention gate.

\subsection{Relevance Scoring and Candidate Pool Construction}
\label{sec:relevance_scoring}

For strategies that rely on explicit evidence ranking (B2, B4, and B5), we first score each evidence unit for relevance to the claim. Given a claim $c$ and the full evidence set $\mathcal{E}$, all units are scored jointly in a single pass: the model assigns each unit $e_i$ a pointwise relevance score $r_i \in [0, 1]$, reflecting how directly the unit relates to verifying the claim. The same scoring criterion is applied uniformly across modalities, so that visual descriptions, ASR segments, and external search results are compared on a common scale. No post-hoc normalization is applied. Units are then sorted by score in descending order to produce a ranked list shared by B2, B4, and B5. Scores are used strictly for ranking and candidate selection; they do not enter the verification stage as features or weights.

For B4 and B5, the ranked list is further refined into a balanced candidate pool of at most 10 units, a budget chosen to balance evidence coverage against prompt length. To prevent single-modality dominance---particularly by ASR segments, which account for over half of all units (Section~\ref{sec:dataset_stats})---the pool is constructed to include at least 2 internal units (V or A) and at least 3 external units (E), with remaining slots filled by the highest-scoring units regardless of modality. The higher minimum for external units reflects the finding that external evidence is critical for factual verdict determination (Section~\ref{sec:ablation}), while internal evidence primarily supports claim--video consistency assessment. This constraint ensures that the sufficiency search operates over both internal and external evidence types.

\subsection{Evidence Selection Strategies}
\label{sec:evidence_selection}

\subsubsection{Baseline Strategies}
B1 (Full Evidence) presents all available units to the verifier without selection, serving as the full-evidence reference. External units are capped at 8 to limit prompt length. B2 (Top-K) selects the $k$ highest-scoring units from the ranked list and passes them directly to the verifier; we evaluate $k = 3, 5, 7$ to trace the relationship between evidence budget and verification performance. B3 (CoT Self-Selection) provides all evidence units and instructs the model to analyze each unit's relationship to the claim, select a subset it deems sufficient, and produce a verdict based on that subset---all within a single inference pass.

\subsubsection{Sufficiency-Driven Selection}
The preceding strategies either use all evidence or select by relevance score, but none explicitly models whether the selected subset is \textit{sufficient} for a confident verdict. We define sufficiency operationally: a selected evidence subset $\mathcal{S}$ is treated as sufficient when the verifier judges it adequate to support a verdict with confidence above a predefined threshold $\tau$. We set $\tau = 0.8$ as a conservative threshold.

B4 (Greedy-Sufficient) operates in three phases on the balanced candidate pool (Section~\ref{sec:relevance_scoring}):

\textbf{Phase 1: Greedy Addition.} Starting from an empty set, units are added one at a time in descending relevance order. After each addition, a sufficiency check queries the model: given the current subset and the claim, is the evidence sufficient to reach a confident verdict? If the model returns \textit{sufficient} with confidence $\geq \tau$, addition stops. If the entire candidate pool is exhausted without reaching sufficiency, all candidates proceed to the next phase.

\textbf{Phase 2: Redundancy Pruning.} Each unit in the selected set is tested for removal: if the remaining units still satisfy the sufficiency criterion, the unit is permanently discarded as redundant. This phase yields a compact sufficient subset under the greedy pruning procedure---no single remaining unit can be removed without falling below the sufficiency threshold, though the result is not guaranteed to be globally minimal.

\textbf{Phase 3: Final Verification.} The pruned subset is passed to the two-layer verification framework (Section~\ref{sec:two_layer_verification}) to produce the final verdict.

B5 (Greedy + Abstention) follows the same procedure as B4, with one key modification: if Phase 1 exhausts the candidate pool without reaching sufficiency, the system outputs \textit{insufficient} as its verdict and skips Phases 2 and 3. This mechanism allows the system to explicitly acknowledge when the available evidence is inadequate, rather than forcing a verdict from an insufficient evidence base.

\subsection{Two-Layer Verification}
\label{sec:two_layer_verification}

All strategies produce their final verdict through a shared two-layer verification framework.

\textbf{Layer 1: Internal-Evidence Grounding.} Using only the selected internal evidence units (V and A), the model assesses whether the claim is compatible with the video-derived evidence representation. The output is one of three labels: \textit{consistent}, \textit{inconsistent}, or \textit{unclear}. Because both the claim and the internal evidence are derived from the same underlying video (Section~\ref{sec:data_collection}), this layer is not intended as a strong discriminative module on its own; rather, it encourages the verifier to anchor its reasoning in video-internal content before consulting external sources, and provides an interpretable intermediate signal about internal alignment.

\textbf{Layer 2: Factual Verdict.} Using the full selected subset $\mathcal{S}$, including external evidence (E units), the model determines the factual status of the claim. The output is \textit{supported}, \textit{refuted}, or \textit{insufficient}, accompanied by a confidence score and reasoning. The final verdict is always determined by this layer.

The two layers are executed jointly in a single inference call. Decoupling internal grounding from factual verification reflects the observation that internal compatibility does not imply factual truth---a video may faithfully depict content described by the claim while the underlying event is fabricated or misattributed---and that external evidence is necessary to bridge this gap.

One exception to this flow arises in B5: when the selection stage fails to reach sufficiency (Section~\ref{sec:evidence_selection}), the system outputs \textit{insufficient} directly, bypassing the two-layer verification entirely.

\section{Experiments}
\subsection{Experimental Setup}
\label{sec:experimental_setup}

\textbf{Model and configuration.} All five strategies use the same backbone model---Claude Sonnet 4 (claude-sonnet-4-20250514, Anthropic)---for relevance scoring, sufficiency checking, and verdict prediction. The temperature is set to 0.1 to reduce output variance, and the maximum generation length is 4,096 tokens. No strategy-specific tuning is applied: prompts differ only in the evidence subset provided and the task instruction (scoring, sufficiency check, or verification). In all prompts, each evidence unit is truncated to its first 150 characters (100 characters for relevance scoring), for every modality and strategy. Section~\ref{sec:crossmodel} repeats the pipeline with two further verifiers.

\textbf{Evaluation metrics.} We report five metrics. \textit{Macro-F1} is the primary metric, computed as the unweighted average of per-class F1 scores across the three verdict categories (supported, refuted, insufficient). Because the label distribution is imbalanced (Section~\ref{sec:data_collection}), Macro-F1 gives equal weight to each category regardless of its sample count, penalizing methods that sacrifice minority-class performance for majority-class gains. \textit{Accuracy} is the fraction of correctly classified samples. \textit{Abstention Precision} is the precision of the \textit{insufficient} prediction: among all samples predicted as insufficient, the fraction that are truly insufficient. \textit{Abstention Recall} is the corresponding recall: among truly insufficient samples, the fraction correctly predicted. \textit{Overconfidence Rate} is the complement of Abstention Recall ($1 - \text{Abstention Recall}$), measuring the proportion of truly insufficient samples that receive a non-insufficient verdict---a proxy for the system's tendency to fail to abstain when evidence is genuinely inadequate.

We additionally report \textit{average evidence units per sample} for each strategy, as evidence efficiency is a central concern of this work.

\subsection{Main Results}
\label{sec:main_results}

Table~\ref{tab:main_results} presents the main results across all five strategies.

\begin{table}[htbp]
\centering
\caption{Main results (195 samples, Claude Sonnet 4). The lower block applies confidence-threshold abstention to three base strategies: any committed verdict with confidence below $\tau = 0.8$ (the threshold B4 and B5 use for the sufficiency check) is converted to \textit{insufficient}. Bold marks the highest Macro-F1 among efficient methods ($\leq$10 units).}
\label{tab:main_results}
\begin{tabular}{lcccccc}
\toprule
Method & Macro-F1 & Accuracy & Avg Units & Abst. Prec. & Abst. Recall & OverConf. \\
\midrule
B1 Full & 0.518 & 56.9\% & 27.7 & 0.500 & 0.254 & 0.746 \\
B2 Top-3 & 0.466 & 49.7\% & 3.0 & 0.378 & 0.525 & 0.475 \\
B2 Top-5 & 0.459 & 50.3\% & 5.0 & 0.370 & 0.339 & 0.661 \\
B2 Top-7 & 0.460 & 51.3\% & 7.0 & 0.350 & 0.237 & 0.763 \\
B3 CoT & 0.453 & 57.4\% & 9.4 & 1.000 & 0.051 & 0.949 \\
B4 Greedy & 0.469 & 52.3\% & 4.1 & 0.333 & 0.203 & 0.797 \\
\textbf{B5 Abstention} & \textbf{0.510} & 54.4\% & 4.5 & 0.377 & 0.390 & 0.610 \\
\midrule
\multicolumn{7}{l}{\textit{Confidence-threshold abstention ($\tau = 0.8$)}} \\
B1 Full + $\tau$ & 0.525 & 57.4\% & 27.7 & 0.516 & 0.271 & 0.729 \\
B2 Top-5 + $\tau$ & 0.464 & 50.8\% & 5.0 & 0.382 & 0.356 & 0.644 \\
B4 Greedy + $\tau$ & 0.476 & 52.8\% & 4.1 & 0.351 & 0.220 & 0.780 \\
\bottomrule
\end{tabular}
\end{table}

We assess differences with a paired bootstrap over the 195 videos (2,000 resamples); full results, including the cross-model backbones, are in Appendix~\ref{app:significance}.

\textbf{Sufficiency-driven selection with abstention matches full-evidence performance at a fraction of the cost.} B5, our proposed method, achieves a Macro-F1 of 0.510 using an average of 4.5 evidence units---approximately 16\% of the full evidence set---recovering 98.5\% of the full-evidence baseline B1's performance (0.518 with 27.7 units). The difference between B5 and B1 is not statistically significant ($\Delta$Macro-F1 = $-$0.008, 95\% CI [$-$0.08, +0.06]): B1's slight advantage comes at over 6$\times$ the evidence cost. Among efficient methods (B2 through B5), B5 attains the highest Macro-F1, ahead of B2--B4 by 0.04 to 0.06. These margins are consistent in sign, but at $n{=}195$ none of them is statistically significant.

\textbf{Fixed-budget relevance selection offers no principled budget.} The B2 Top-K variants cluster below 0.47 Macro-F1 for every $k$ (Top-3 0.466, Top-5 0.459, Top-7 0.460), while their abstention behavior shifts sharply with $k$ (Abstention Recall 0.525 $\to$ 0.237), leaving no principled way to choose $k$. We attribute this to relevance to the claim not guaranteeing sufficiency for verification---a highly relevant unit may duplicate information already present in the selected set, while a moderately relevant unit may provide the missing factual link needed for a verdict. B5 eliminates the need to pre-specify $k$ by dynamically determining how much evidence is enough on a per-sample basis.

\textbf{CoT self-selection exhibits systematic overcommitment.} B3 achieves the highest accuracy among the base strategies (57.4\%) but the lowest Macro-F1 (0.453), exposing a strong tendency toward unjustified non-abstention. B3 predicts \textit{insufficient} for only 3 out of 195 samples, yielding an Overconfidence Rate of 94.9\%---the highest among all methods. It correctly classifies 84/97 supported cases (86.6\%) but only 3/59 insufficient cases (5.1\%). When the model self-selects evidence and produces a verdict in a single pass, it consistently overestimates the adequacy of its evidence, committing to a verdict even when the evidence is genuinely inadequate. This finding underscores the need for an explicit abstention mechanism rather than leaving abstention to the model's own judgment during verdict generation.

\textbf{Explicit abstention improves recognition of insufficient cases.} Comparing B4 and B5 isolates the contribution of the abstention gate. B5 improves Macro-F1 from 0.469 to 0.510 (+0.041, not significant at $n{=}195$), with the gain concentrated in the insufficient class: insufficient-class F1 rises from 0.253 to 0.383 (+0.131, 95\% CI [+0.03, +0.23], $p{=}0.010$), and Abstention Recall nearly doubles, from 0.203 to 0.390 ($p{<}0.001$). Without abstention, B4 is forced to commit to a verdict even when the candidate pool fails to reach sufficiency, producing overconfident errors on difficult samples. The abstention mechanism converts these potential misclassifications into calibrated \textit{insufficient} outputs.

\textbf{Confidence thresholding is a competitive alternative abstention mechanism.} A simpler way to abstain is to threshold the verifier's self-reported confidence. With $\tau = 0.8$, the threshold B4 and B5 already use for the sufficiency check, thresholding raises Macro-F1 for every base strategy (lower block of Table~\ref{tab:main_results}): B1 to 0.525, B2 Top-5 to 0.464, and B4 to 0.476. B5 is statistically indistinguishable from thresholded full evidence ($\Delta$Macro-F1 = $-$0.015, 95\% CI [$-$0.09, +0.05]) while using 16\% of the evidence, and its insufficient-class F1 is significantly higher than that of thresholded B4 (+0.113, $p{=}0.028$). Tuning $\tau$ by 5-fold cross-validation does not change this picture: B1 + $\tau$ reaches 0.534 $\pm$ 0.014 and B4 + $\tau$ 0.471 $\pm$ 0.013 over 20 random splits, and neither differs significantly from B5. Explicit abstention therefore matters more than its specific form---both the sufficiency gate and a confidence threshold counter the overcommitment of B3 and B4. What distinguishes B5 is that it reaches full-evidence performance with a small evidence subset that is itself judged sufficient for the verdict, so each abstention is tied to an explicit, inspectable failure to reach sufficiency.

\textbf{Per-class breakdown.} B2 Top-3 achieves the highest insufficient accuracy (52.5\%) but at the cost of much lower supported and refuted accuracy (54.6\% / 33.3\%). B5 reaches 39.0\% on insufficient cases while maintaining stronger supported and refuted accuracy (68.0\% / 43.6\%), yielding the most balanced profile.

\subsection{Ablation Studies}
\label{sec:ablation}

To isolate the contribution of individual components and evidence sources, we conduct five ablation experiments (Table~\ref{tab:ablation}). The abstention gate is ablated by comparing B5 with B4, which is identical except that it never abstains. The evidence-source and pruning ablations each remove one component from B4. We run them without the abstention gate because the gate outputs \textit{insufficient} whenever the candidate pool is exhausted: with a weakened evidence pool, it would turn evidence loss into abstention by construction, and the measured effect would reflect the gate rather than the verifier's verdicts. $\Delta$F1 is therefore reported relative to B5 for the abstention ablation and relative to B4 for all others, with 95\% confidence intervals from a paired bootstrap over the 195 videos (2,000 resamples).

\begin{table}[htbp]
\centering
\small
\caption{Ablation study results. The abstention gate is ablated by comparing B5 with B4; all other components are ablated on B4 (the selection pipeline without the abstention gate), and their $\Delta$F1 is relative to B4. Brackets give 95\% paired-bootstrap confidence intervals.}
\label{tab:ablation}
\begin{tabular}{lcccc}
\toprule
Setting & Macro-F1 & Acc. & Units & $\Delta$F1 [95\% CI] \\
\midrule
\textbf{B5 (proposed)} & \textbf{0.510} & 54.4\% & 4.5 & --- \\
w/o Abstention (= B4) & 0.469 & 52.3\% & 4.1 & $-$0.041 [$-$0.10, +0.02] \\
\midrule
\multicolumn{5}{l}{\textit{Component ablations on B4 ($\Delta$F1 relative to B4)}} \\
w/o External & 0.273 & 32.3\% & 7.3 & $-$0.196 [$-$0.28, $-$0.11] \\
w/o V+A (E-only) & 0.484 & 53.3\% & 3.5 & +0.016 [$-$0.05, +0.08] \\
w/o ASR & 0.465 & 51.3\% & 4.6 & $-$0.004 [$-$0.07, +0.06] \\
w/o Pruning & 0.477 & 52.8\% & 5.4 & +0.008 [$-$0.04, +0.05] \\
\bottomrule
\end{tabular}
\end{table}

\textbf{Abstention improves insufficient-case recognition.} Removing the abstention gate lowers Macro-F1 from 0.510 to 0.469. This overall difference is not statistically significant at $n{=}195$, but the effect is concentrated where the gate is designed to act: insufficient-class F1 and Abstention Recall both drop significantly (Section~\ref{sec:main_results}).

\textbf{External evidence is indispensable ($\Delta$F1 = $-$0.196).} Removing external web-retrieved evidence causes the most severe performance collapse across all experiments, and the only significant one among the component ablations. Macro-F1 drops from 0.469 to 0.273, and the verifier returns \textit{insufficient} for 153 of 195 samples (78.5\%) even though B4 has no abstention gate. The Layer~1 outputs show why: with internal evidence alone, the verifier judges 134 claims (68.7\%) \textit{consistent} with the video, yet still cannot determine whether they are true. Average evidence consumption rises from 4.1 to 7.3 units, as the search frequently exhausts the candidate pool without reaching the sufficiency threshold. Claim--video consistency is therefore no substitute for external corroboration, which is the single most critical component for factual verdict determination in short-video verification.

\textbf{Internal evidence grounds the verdict rather than determining it ($\Delta$F1 = +0.016, n.s.).} Removing visual keyframes and ASR transcripts (E-only) leaves Macro-F1 statistically unchanged (0.484 vs.\ 0.469), indicating that external evidence carries the dominant signal for verdict determination. Verdict accuracy, however, does not capture the role of internal evidence: in the E-only setting, all 195 samples receive \textit{unclear} as their Layer~1 consistency assessment, whereas with internal evidence available, B4 reaches a definite consistency judgment for 96 samples (49.2\%). Without internal evidence, the verdict is reached with no link to the video's content. Internal evidence is thus what makes the two-layer framework interpretable: it anchors the verifier's reasoning in video-derived content, which is a prerequisite for recognizing videos that are internally consistent with their claims yet factually misleading.

Removing ASR transcripts alone likewise leaves Macro-F1 unchanged ($\Delta$F1 = $-$0.004, n.s.). ASR segments help establish what the video \textit{claims} but rarely provide independent factual corroboration, so their contribution lies in internal grounding rather than factual resolution.

\textbf{Pruning improves efficiency without loss of accuracy ($\Delta$F1 = +0.008, n.s.).} Skipping the redundancy pruning phase leaves Macro-F1 statistically unchanged (0.477 vs.\ 0.469) but increases average evidence from 4.1 to 5.4 units. Pruning thus removes 24\% of the selected units while preserving verdict quality: the discarded units are redundant, since the remaining subset still satisfies the sufficiency criterion. This is precisely the property that minimal sufficient evidence identification targets---a compact subset that supports the same verdict as a larger one.

\subsection{Evidence Efficiency Analysis}
\label{sec:efficiency}

\begin{figure}[t]
\centering
\includegraphics[width=0.65\textwidth]{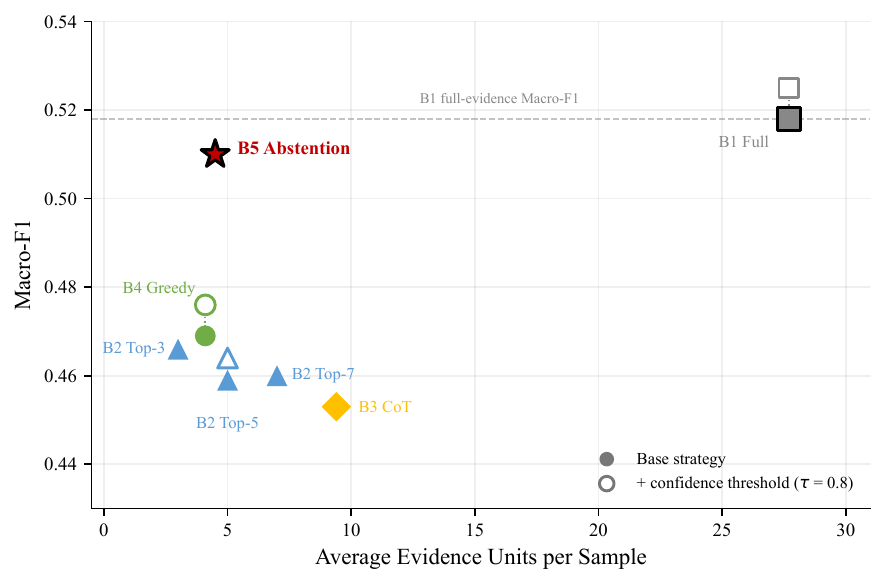}
\caption{Macro-F1 vs.\ average evidence units per sample across all strategies. B5 (star) achieves near-full-evidence performance (98.5\% of B1) using only 16\% of the evidence. Hollow markers apply confidence-threshold abstention ($\tau = 0.8$) to B1, B2 Top-5, and B4. The dashed line marks B1's full-evidence Macro-F1.}
\label{fig:efficiency_tradeoff}
\end{figure}

A central claim of this work is that a small, sufficiency-aware evidence subset can recover most of the performance achieved by full evidence. Figure~\ref{fig:efficiency_tradeoff} illustrates this trade-off across all strategies.

B5 operates at a favorable point on the efficiency--performance frontier: it uses 4.5 units (16\% of B1's 27.7) while recovering 98.5\% of B1's Macro-F1. By contrast, B2 Top-7 uses 7.0 units (25\%) but achieves only 88.8\% of B1's performance, and B3 uses 9.4 units (34\%) while achieving only 87.5\%. The sufficiency-driven approach dynamically allocates evidence per sample---easy cases are resolved with 1--2 units, while genuinely difficult cases consume the full candidate pool---rather than applying a uniform budget across all samples.

The ablation results reinforce this picture: removing external evidence forces the pipeline to consume 7.3 units on average while achieving only 52.7\% of B1's Macro-F1, demonstrating that evidence quantity cannot compensate for the absence of the right evidence type. Conversely, removing pruning increases evidence by 31\% (4.1 $\to$ 5.4) with no significant change in Macro-F1, confirming that the units removed by pruning are redundant: compactness is obtained without sacrificing accuracy.

\subsection{Human Validation of Evidence Selection}
\label{sec:human_validation}

The preceding sections evaluate evidence selection through verdict accuracy, but a natural question remains: does B5 select the \emph{right} evidence? We address this by comparing B5's automatically selected evidence sets against the human annotations described in Section~\ref{sec:human_annotation}. Table~\ref{tab:human_validation} summarizes the comparison across all 195 samples. The high level of overlap suggests that B5's sufficiency-driven selection is broadly consistent with human judgment, though the two are not identical because multiple compact evidence subsets can support the same verdict.

\begin{table}[htbp]
\centering
\caption{Comparison of B5's evidence selection against human annotations (195 samples).}
\label{tab:human_validation}
\begin{tabular}{lc}
\toprule
Metric & Value \\
\midrule
Avg Jaccard similarity & 0.882 \\
Avg Precision (B5 $\cap$ human / B5) & 0.899 \\
Avg Recall (B5 $\cap$ human / human) & 0.970 \\
Exact match rate & 71.8\% \\
Human $\subseteq$ B5 rate & 92.8\% \\
Avg human set size & 3.74 \\
Avg B5 set size & 4.46 \\
\bottomrule
\end{tabular}
\end{table}

\textbf{B5 shows substantial overlap with human annotations.} The average Jaccard similarity between B5's selected sets and human annotations is 0.882, with 71.8\% of samples achieving an exact match. B5's recall with respect to the human annotations is 0.970, indicating that it recovers most of the evidence units identified by human annotators, while its precision of 0.899 suggests that the automated method retains only a small number of additional units. The size difference is modest: B5 selects 4.46 units on average versus 3.74 for human annotators.

\textbf{Human annotations also highlight the importance of external evidence.} The modality composition of human-annotated sets is 51.6\% external (E), 32.5\% ASR (A), and 15.9\% visual (V). This distribution closely mirrors the ablation findings in Section~\ref{sec:ablation}: external evidence is the most frequently selected modality in both automated and human selections, reinforcing its indispensable role in factual verdict determination.

\subsection{Cross-Model Generalization}
\label{sec:crossmodel}

To test whether the findings depend on the verifier, we rerun the full pipeline---relevance scoring, sufficiency search, and two-layer verification---with GPT-5.5 (closed-weight, different provider) and Qwen2.5-72B-Instruct (open-weight) on the identical frozen evidence pool: the same 5,510 units, including the Claude-generated V descriptions. Because the verifier now differs from the description model, this also tests whether the results rely on one model both producing and consuming the visual descriptions. Qwen covers 193 of the 195 videos; the provider's content-safety filter declined two samples about disasters and violent incidents. Table~\ref{tab:crossmodel} summarizes the results; full results and significance tests are in Appendices~\ref{app:crossmodel} and~\ref{app:significance}.

\begin{table}[htbp]
\centering
\small
\setlength{\tabcolsep}{3.5pt}
\caption{Cross-model results on the identical frozen evidence pool. B5/B1: share of B1's Macro-F1 recovered by B5. Units: B5's average evidence units (share of B1's). Ins-F1: insufficient-class F1.}
\label{tab:crossmodel}
\begin{tabular}{lccccc}
\toprule
Verifier & B1 F1 & B5 F1 & B5/B1 & Units & Ins-F1 B4$\to$B5 \\
\midrule
Claude Sonnet 4 & 0.518 & 0.510 & 98.5\% & 4.5 (16\%) & 0.253$\to$0.383 \\
GPT-5.5 & 0.586 & 0.559 & 95.3\% & 4.9 (18\%) & 0.407$\to$0.508 \\
Qwen2.5-72B & 0.580 & 0.489 & 84.3\% & 2.4 (9\%) & 0.413$\to$0.423 \\
\bottomrule
\end{tabular}
\end{table}

Full-evidence Macro-F1 lies in the same band for all three verifiers (0.52--0.59), so the absolute level tracks verifier capability rather than a property of one model. The efficiency result replicates on GPT-5.5: B5 recovers 95.3\% of B1's Macro-F1 with 18\% of the evidence, with no significant difference from B1, and again improves insufficient-class F1 over B4 (+0.100, $p{=}0.025$). On Qwen, B5 recovers 84.3\% of B1's Macro-F1 with 9\% of the evidence but is significantly below both B1 and Top-5, and its abstention gain is small. Qwen is a coarse sufficiency judge: it ends with a single evidence unit on 60\% of samples, compared with 28\% for GPT-5.5 and 17\% for Claude, so its small evidence sets reflect premature stopping rather than genuine minimality. Sufficiency-driven selection therefore transfers across providers, but its benefit depends on the quality of the verifier's sufficiency judgments.

\subsection{Robustness and Error Analysis}
\label{sec:robustness}

\textbf{Textualized vs.\ raw visual evidence.} Because V-units are text descriptions, we test whether textualization discards verification-relevant information. On a stratified subset of 60 videos (20 per verdict class), we rerun B1 with the same verifier (Claude Sonnet 4), replacing each V-unit with its keyframe image and leaving ASR and external evidence unchanged, so that only the form of the visual channel differs. Raw keyframes do not improve verification: Macro-F1 is 0.561 with frames and 0.585 with descriptions ($\Delta$Macro-F1 = $-$0.024, 95\% CI [$-$0.14, +0.09], $p{=}0.69$), and the two settings agree on 78.3\% of verdicts. Within the current claim-verification setting, the concise descriptions thus preserve most of the visual information the verifier uses. This control isolates information loss in the visual channel; it does not cover reasoning over motion or temporal dynamics, which our text-based selection does not model (Section~\ref{sec:limitations}).

\textbf{Speech language and evidence quantity.} We classify a video as English-speech when CJK characters make up at most 10\% of its ASR transcript. B5 performs similarly on Chinese-speech videos (Macro-F1 0.510, $n{=}155$) and English-speech videos (0.500, $n{=}40$), as does B1 (0.509 and 0.540); the English group is small, so these figures are indicative only. Insufficiency is also not an artifact of evidence quantity: the average number of available units is nearly identical across verdict classes (28.1 supported, 28.3 refuted, 28.5 insufficient).

\textbf{Error analysis.} B5 misclassifies 89 of the 195 videos, and most errors involve the insufficient class. The largest group is missed abstention (36): insufficient cases that receive a committed verdict, 30 of them \textit{supported}, after B5 judges a small set (3.6 units on average) sufficient. Over-abstention on supported claims accounts for 23 errors and refuted claims labeled \textit{insufficient} for 15; in 22 of these 38 cases, the search exhausts the candidate pool without reaching sufficiency, and in the rest the final verifier itself returns \textit{insufficient}. Only 15 errors confuse supported and refuted. Refuted-to-insufficient errors include claims that mix real and fabricated elements, which require synthesizing several partially contradicting units into a refutation (case study in Appendix~\ref{app:case_study}). The dominant failure is therefore misjudged sufficiency rather than polarity: the verifier accepts inadequate evidence too early on some samples and fails to find decisive evidence within the pool on others.

\section{Discussion}
\subsection{The Asymmetric Roles of Internal and External Evidence}
\label{sec:discuss_roles}

Our ablation results reveal a striking asymmetry in how different evidence modalities contribute to verification. Removing external evidence collapses Macro-F1 by 0.196, the largest single-component effect, whereas removing internal evidence (V+A) leaves it statistically unchanged. At first glance, this suggests that internal evidence is nearly dispensable, but this interpretation is misleading.

In the E-only setting, every sample receives \textit{unclear} as its Layer~1 consistency assessment. The system can still reach correct verdicts in many cases, but it does so without any grounding in the video's actual content---it effectively performs text-based fact-checking using web-retrieved evidence against a textual claim, bypassing the video entirely. The two-layer framework therefore degenerates into a single-layer system, losing the intermediate signal that connects the verdict to the video.

Internal evidence thus serves a qualitatively different function from external evidence: it does not primarily determine \textit{what} the verdict is, but rather \textit{why} the verdict is connected to the video. Visual keyframes and speech transcripts anchor the verifier's reasoning in video-derived content, in principle helping the system remain grounded in cases where a video is internally consistent with its claim yet factually misleading---such as repurposed footage or out-of-context clips. This division of labor validates the two-layer design: Layer~1 (internal grounding) and Layer~2 (factual determination) address complementary aspects of video verification that cannot be collapsed into a single step without losing interpretability.

\subsection{External Evidence Dominance and Claim-Based Retrieval}
\label{sec:discuss_retrieval}

Claim-based retrieval is a standard component of evidence-based fact-checking~\cite{schlichtkrull2023averitec}, not a methodological shortcoming: verifying a video's factual assertion against external sources is precisely what fact-checking requires. However, the E-only ablation shows that removing internal evidence eliminates all consistency signal, confirming that internal evidence provides grounding that external retrieval cannot supply. Future work could explore video-grounded retrieval---using keyframe content or ASR segments to generate supplementary queries---to reduce dependence on the claim as the sole retrieval anchor.

\subsection{The Necessity of Explicit Abstention}
\label{sec:discuss_abstention}

Our results consistently show that systems without explicit abstention are overconfident on genuinely insufficient cases. B5 mitigates this failure mode by converting unresolved selections into \textit{insufficient} outputs rather than forcing a verdict. A confidence threshold on the verifier's output achieves a comparable effect (Section~\ref{sec:main_results}), which suggests that the essential requirement is an explicit abstention decision rather than a particular mechanism; the sufficiency gate additionally grounds each abstention in a specific evidence set that was checked and found inadequate. This is especially important in high-stakes verification settings, where calibrated uncertainty is preferable to confident error.

\subsection{Limitations}
\label{sec:limitations}

\textbf{Sample size.} MiniVer-V contains 195 videos with 5,510 evidence units. The unit-level decomposition supports controlled comparisons of selection strategies within the same videos, but the number of videos limits statistical power: differences of a few Macro-F1 points between strategies are not significant at this size (Section~\ref{sec:main_results}). It also limits the diversity of topics, platforms, claim types, and evidence configurations. Scaling MiniVer-V is a priority for future work.

\textbf{Verifier and description models.} We evaluate three verifiers (Claude Sonnet 4, GPT-5.5, and Qwen2.5-72B), but all V descriptions are generated by Claude Sonnet 4; testing other description models remains future work. The benefit of sufficiency-driven selection also depends on the verifier's sufficiency judgments, as the weaker results with Qwen show (Section~\ref{sec:crossmodel}).

\textbf{Textualized evidence.} All evidence is consumed as text. The raw-keyframe control (Section~\ref{sec:robustness}) suggests little information loss for claim verification, but the framework does not model motion or temporal dynamics, and MiniVer-V evaluates evidence selection over textualized units rather than full video-level multimodal reasoning.

\textbf{Retrieval of existing fact-checks.} External evidence is retrieved with the claim text, and the second query round adds fact-checking keywords. The results can therefore include fact-checking articles about the same claim, which may state the verdict directly. We did not filter such articles or restrict retrieval to dates before a claim was checked, so the dominant role of external evidence (Section~\ref{sec:ablation}) may partly reflect access to existing fact-checks rather than independent corroboration. Filtering fact-checking sources or applying temporal cutoffs is a natural next step.

\textbf{Consistency evaluation.} Because claims in MiniVer-V are written by annotators based on the video content, most samples are expected to be internally compatible with the video representation. As a result, the benchmark provides limited coverage of naturally inconsistent claim--video pairs (e.g., repurposed footage paired with fabricated claims), which limits evaluation of Layer~1's discriminative power. Introducing adversarial or synthetic inconsistent samples would provide a more rigorous test of the internal grounding layer.

\textbf{Language coverage.} All claims are written in Chinese, and only 40 videos contain mainly English speech. Performance on other languages and in cross-lingual verification settings remains untested.

\section{Conclusion}
\label{sec:conclusion}
We introduced minimal sufficient evidence set identification as a new task for short-video verification and presented MiniVer-V, a benchmark designed to study evidence sufficiency rather than evidence relevance alone. Across five evidence selection strategies, our sufficiency-driven method with explicit abstention recovers 98.5\% of full-evidence performance using only 16\% of the evidence, while improving recognition of genuinely insufficient cases. Human annotation further shows substantial overlap between B5's selected evidence sets and human annotations (Jaccard 0.882). These findings indicate that evidence-efficient, calibrated short-video verification depends not on presenting more evidence, but on selecting enough evidence, grounding predictions in video-internal content, and abstaining when that threshold is not met. Future work includes scaling MiniVer-V to a larger and more diverse sample set, testing description models beyond Claude, and exploring video-grounded retrieval strategies.

\section*{Ethics and Data Availability}

\textbf{Ethics statement.} The videos in MiniVer-V were collected from publicly available posts on social media platforms and are used solely for non-commercial research. All annotations---claims, verdict labels, and reference evidence sets---were produced by the three authors; no crowdsourcing platform was used. Because the dataset contains news and public-affairs content, we reviewed the samples to ensure that no personally identifiable information is disclosed beyond what is already public in the original posts. The benchmark is intended to support fact-checking research, not to target individuals or communities.

\textbf{Data availability.} We plan to release the MiniVer-V evidence units, verdict annotations, human-annotated reference evidence sets, and the code for evidence extraction, selection, and evaluation. Raw video files will not be redistributed; we will provide source references for obtaining the videos, subject to each platform's terms of service.

\bibliographystyle{plainnat}
\bibliography{ref}

@inproceedings{qi2023fakesv,
  title     = {{FakeSV}: A Multimodal Benchmark with Rich Social Context for Fake News Detection on Short Video Platforms},
  author    = {Qi, Peng and Bu, Yuyan and Cao, Juan and Ji, Wei and Shui, Ruihao and Xiao, Junbin and Wang, Danding and Chua, Tat-Seng},
  booktitle = {Proceedings of the AAAI Conference on Artificial Intelligence},
  volume    = {37},
  pages     = {14444--14452},
  year      = {2023},
  doi       = {10.1609/aaai.v37i12.26689}
}

@inproceedings{bu2024fakingrecipe,
  title     = {{FakingRecipe}: Detecting Fake News on Short Video Platforms from the Perspective of Creative Process},
  author    = {Bu, Yuyan and Sheng, Qiang and Cao, Juan and Qi, Peng and Wang, Danding and Li, Jintao},
  booktitle = {Proceedings of the 32nd ACM International Conference on Multimedia},
  pages     = {1351--1360},
  publisher = {ACM},
  year      = {2024},
  doi       = {10.1145/3664647.3680663}
}

@inproceedings{shao2023dgm4,
  title     = {Detecting and Grounding Multi-Modal Media Manipulation},
  author    = {Shao, Rui and Wu, Tianxing and Liu, Ziwei},
  booktitle = {Proceedings of the IEEE/CVF Conference on Computer Vision and Pattern Recognition (CVPR)},
  pages     = {6904--6913},
  year      = {2023},
  doi       = {10.1109/CVPR52729.2023.00667}
}

@inproceedings{thorne2018fever,
  title     = {{FEVER}: a Large-scale Dataset for Fact Extraction and {VER}ification},
  author    = {Thorne, James and Vlachos, Andreas and Christodoulopoulos, Christos and Mittal, Arpit},
  booktitle = {Proceedings of the 2018 Conference of the North {A}merican Chapter of the Association for Computational Linguistics: Human Language Technologies, Volume 1 (Long Papers)},
  pages     = {809--819},
  address   = {New Orleans, Louisiana},
  publisher = {Association for Computational Linguistics},
  year      = {2018},
  doi       = {10.18653/v1/N18-1074}
}

@inproceedings{yao2023end,
  title     = {End-to-End Multimodal Fact-Checking and Explanation Generation: A Challenging Dataset and Models},
  author    = {Yao, Barry Menglong and Shah, Aditya and Sun, Lichao and Cho, Jin-Hee and Huang, Lifu},
  booktitle = {Proceedings of the 46th International ACM SIGIR Conference on Research and Development in Information Retrieval},
  pages     = {2733--2743},
  publisher = {ACM},
  year      = {2023},
  doi       = {10.1145/3539618.3591879}
}

@article{atanasova2022fact,
  title   = {Fact Checking with Insufficient Evidence},
  author  = {Atanasova, Pepa and Simonsen, Jakob Grue and Lioma, Christina and Augenstein, Isabelle},
  journal = {Transactions of the Association for Computational Linguistics},
  volume  = {10},
  pages   = {746--763},
  year    = {2022}
}

@inproceedings{pan2023fact,
  title     = {Fact-Checking Complex Claims with Program-Guided Reasoning},
  author    = {Pan, Liangming and Wu, Xiaobao and Lu, Xinyuan and Luu, Anh Tuan and Wang, William Yang and Kan, Min-Yen and Nakov, Preslav},
  booktitle = {Proceedings of the 61st Annual Meeting of the Association for Computational Linguistics (Volume 1: Long Papers)},
  pages     = {6981--7004},
  publisher = {Association for Computational Linguistics},
  year      = {2023},
  doi       = {10.18653/v1/2023.acl-long.386}
}

@inproceedings{zhang2025fakevv,
  title     = {{Fact-R1}: Towards Explainable Video Misinformation Detection with Deep Reasoning},
  author    = {Zhang, Fanrui and Li, Dian and Zhang, Qiang and Chen, Jun and Liu, Gang and Lin, Junxiong and Yan, Jiahong and Liu, Jiawei and Zha, Zheng-Jun},
  booktitle = {Advances in Neural Information Processing Systems},
  volume    = {38},
  publisher = {Curran Associates, Inc.},
  year      = {2025},
  doi       = {10.52202/085713-5438}
}

@inproceedings{zeng2026wildfakebench,
  title     = {From Manipulation to Mistrust: Explaining Diverse Micro-Video Misinformation for Robust Debunking in the Wild},
  author    = {Zeng, Zhi and Yang, Yifei and Wu, Jiaying and Zhang, Xulang and Kong, Xiangzheng and Wan, Herun and Ma, Zihan and Luo, Minnan},
  booktitle = {Proceedings of the ACM Web Conference 2026},
  pages     = {7621--7632},
  publisher = {ACM},
  year      = {2026},
  doi       = {10.1145/3774904.3792721}
}

@inproceedings{schlichtkrull2023averitec,
  title     = {{AVeriTeC}: A Dataset for Real-world Claim Verification with Evidence from the Web},
  author    = {Schlichtkrull, Michael and Guo, Zhijiang and Vlachos, Andreas},
  booktitle = {Advances in Neural Information Processing Systems, Datasets and Benchmarks Track},
  volume    = {36},
  pages     = {65128--65167},
  publisher = {Curran Associates, Inc.},
  year      = {2023}
}

@inproceedings{braun2025defame,
  title     = {{DEFAME}: Dynamic Evidence-based {FA}ct-checking with Multimodal Experts},
  author    = {Braun, Tobias and Rothermel, Mark and Rohrbach, Marcus and Rohrbach, Anna},
  booktitle = {Proceedings of the 42nd International Conference on Machine Learning},
  series    = {Proceedings of Machine Learning Research},
  volume    = {267},
  pages     = {5383--5417},
  publisher = {PMLR},
  year      = {2025}
}

@article{guo2022survey,
  title   = {A Survey on Automated Fact-Checking},
  author  = {Guo, Zhijiang and Schlichtkrull, Michael and Vlachos, Andreas},
  journal = {Transactions of the Association for Computational Linguistics},
  volume  = {10},
  pages   = {178--206},
  year    = {2022}
}

@inproceedings{roberts2024scifibench,
  title     = {{SciFIBench}: Benchmarking Large Multimodal Models for Scientific Figure Interpretation},
  author    = {Roberts, Jonathan and Han, Kai and Houlsby, Neil and Albanie, Samuel},
  booktitle = {Advances in Neural Information Processing Systems, Datasets and Benchmarks Track},
  volume    = {37},
  pages     = {18695--18728},
  publisher = {Curran Associates, Inc.},
  year      = {2024}
}

@inproceedings{liu2020kgat,
  title     = {Fine-grained Fact Verification with Kernel Graph Attention Network},
  author    = {Liu, Zhenghao and Xiong, Chenyan and Sun, Maosong and Liu, Zhiyuan},
  booktitle = {Proceedings of the Annual Meeting of the Association for Computational Linguistics},
  pages     = {7342--7351},
  year      = {2020}
}

@inproceedings{zheng2024evidence,
  title     = {Evidence Retrieval is almost All You Need for Fact Verification},
  author    = {Zheng, Liwen and Li, Chaozhuo and Zhang, Xi and Shang, Yu-Ming and Huang, Feiran and Jia, Haoran},
  booktitle = {Findings of the Association for Computational Linguistics: ACL 2024},
  pages     = {9274--9281},
  publisher = {Association for Computational Linguistics},
  year      = {2024},
  doi       = {10.18653/v1/2024.findings-acl.551}
}

@article{wen2025know,
  title   = {Know Your Limits: A Survey of Abstention in Large Language Models},
  author  = {Wen, Bingbing and Yao, Jihan and Feng, Shangbin and Xu, Chenjun and Tsvetkov, Yulia and Howe, Bill and Wang, Lucy Lu},
  journal = {Transactions of the Association for Computational Linguistics},
  volume  = {13},
  pages   = {529--556},
  year    = {2025}
}

@inproceedings{shi2023distracted,
  title     = {Large Language Models Can Be Easily Distracted by Irrelevant Context},
  author    = {Shi, Freda and Chen, Xinyun and Misra, Kanishka and Scales, Nathan and Dohan, David and Chi, Ed H. and Sch{\"a}rli, Nathanael and Zhou, Denny},
  booktitle = {Proceedings of the 40th International Conference on Machine Learning},
  series    = {Proceedings of Machine Learning Research},
  volume    = {202},
  pages     = {31210--31227},
  publisher = {PMLR},
  year      = {2023}
}

\appendix

\section{Prompt Templates}
\label{app:prompts}

All prompts follow the same instruction format and were implemented in Chinese, matching the dominant language of the dataset. We present English translations of the four core prompt templates used across all five baseline strategies.

\subsection{Two-Layer Verification Prompt}
\label{app:prompt_verify}

This prompt is shared by all five strategies (B1--B5) for final verdict prediction. The \texttt{\{claim\}} and \texttt{\{evidence\_text\}} placeholders are filled at runtime.

\begin{quote}
\small
\textit{You are a professional fact-checker. Please make a \textbf{two-layer judgment} based on the following evidence.}

\textbf{Claim:} \texttt{\{claim\}}

\textbf{Evidence:} \texttt{\{evidence\_text\}}

\textbf{Task:} Please perform two layers of analysis:

\textbf{Layer 1: Claim--Video Consistency.}
Using only internal video evidence (V-type visual descriptions + A-type speech transcripts), determine whether the claim is consistent with the video content:
``consistent'' (matches), ``inconsistent'' (contradicts), or ``unclear'' (cannot judge from internal evidence alone).

\textbf{Layer 2: Factual Verdict.}
Using all evidence (including E-type external search results), determine the factual status of the claim:
``supported'' (evidence supports the claim), ``refuted'' (evidence disproves the claim), or ``insufficient'' (evidence is inadequate for a reliable judgment).

\textit{Note: Video consistency does not imply truth (the video may be repurposed footage); video inconsistency does not imply falsehood (the claim may be loosely worded but factually correct). External evidence can overturn preliminary judgments based solely on video content.}

Output in JSON format:\\
{\small\ttfamily \{"consistency": "...", "verdict": "...",\\
\phantom{\{}"confidence": 0.0--1.0, "reasoning": "..."\}\par}
\end{quote}

\subsection{Relevance Scoring Prompt}
\label{app:prompt_relevance}

Used by B2 (Top-K), B4, and B5 to score all evidence units for relevance in a single batch call.

\begin{quote}
\small
\textit{Given a claim and multiple evidence units, rate each unit's relevance to the claim (0--1).}

\textbf{Claim:} \texttt{\{claim\}}

\textbf{Evidence list:} \texttt{\{evidence\_list\}}

Output in JSON format:
\texttt{\{"scores": [\{"unit\_id": "...", "score": 0.0--1.0\}]\}}

\textit{Output JSON only, no other content.}
\end{quote}

\subsection{Chain-of-Thought Selection Prompt}
\label{app:prompt_cot}

Used by B3 (CoT Self-Selection). The model analyzes, selects, and verifies in a single pass.

\begin{quote}
\small
\textit{You are a fact-checking expert. Given a claim and candidate evidence, please:}
\begin{enumerate}
\item \textit{Analyze each evidence unit's relationship to the claim}
\item \textit{Select a subset you consider sufficient for a verdict (considering both internal and external evidence)}
\item \textit{Based on the selected evidence, provide consistency and verdict judgments}
\end{enumerate}

\textbf{Claim:} \texttt{\{claim\}}

\textbf{Candidate evidence:} \texttt{\{evidence\_text\}}

\textbf{Evidence type key:}
V-type (video\_segment): visual keyframe descriptions;
A-type (asr): speech transcripts;
E-type (external): web-retrieved evidence (Google Search)

Output in JSON format:\\
{\small\ttfamily \{"selected\_units": [...], "consistency": "...", "verdict": "...",\\
\phantom{\{}"confidence": 0.0--1.0, "reasoning": "..."\}\par}
\end{quote}

\subsection{Sufficiency Check Prompt}
\label{app:prompt_sufficiency}

Used by B4 and B5 during the greedy addition and pruning phases to assess whether the current evidence subset is sufficient.

\begin{quote}
\small
\textit{Given a claim and a set of evidence, determine whether this evidence set is \textbf{sufficient} to support a verification conclusion.}

\textbf{Claim:} \texttt{\{claim\}}

\textbf{Evidence:} \texttt{\{evidence\_text\}}

\textit{``Sufficient'' means: based solely on this evidence, a rational fact-checker could confidently reach a supported or refuted conclusion.
Note: Sufficiency requires considering both internal video evidence (V-type visual + A-type speech for consistency) and external evidence (E-type for factual verification).}

Output in JSON format:\\
{\small\ttfamily \{"is\_sufficient": true/false, "consistency": "...",\\
\phantom{\{}"verdict\_if\_sufficient": "supported/refuted/null", "confidence": 0.0--1.0,\\
\phantom{\{}"missing\_info": "if insufficient, what information is missing"\}\par}
\end{quote}

\section{Annotation Guidelines and Procedure}
\label{app:annotation}

\subsection{Three-Layer Annotation Scheme}

Our annotation follows a three-layer scheme, each targeting a different aspect of evidence quality:

\textbf{Layer 1 (L1): Evidence Unit Segmentation Quality.}
A random 20\% of processed samples are inspected to verify that (i)~ASR transcript segmentation boundaries are reasonable, (ii)~keyframe sampling captures distinct visual content, and (iii)~external search results are non-degenerate (i.e., not empty or irrelevant). Units that fail inspection are re-extracted or manually corrected.

\textbf{Layer 2 (L2): Evidence--Claim Relation Labels.}
Each evidence unit is labeled with its relation to the claim: \textit{support} (provides evidence for the claim), \textit{refute} (provides evidence against the claim), or \textit{irrelevant} (does not bear on the claim's truth value). This layer is initially populated by the backbone VLM (Claude Sonnet 4) and audited by a human reviewer on a random 20\% subset. These relation labels serve as analysis metadata and are not used as input to any verification strategy.

\textbf{Layer 3 (L3): Minimal Sufficient Evidence Set Annotation.}
Three graduate-student annotators independently select the smallest subset of evidence units they consider sufficient to reach the ground-truth verdict. Annotators are shown the claim, ground-truth verdict, and all evidence units grouped by modality. The reference set for each sample is constructed by retaining units selected by at least two of the three annotators (majority vote).

\subsection{Annotation Quality}

\textbf{Inter-annotator agreement.} For the ground-truth verdict labels (Section~\ref{sec:data_collection}), a second reviewer independently annotates a random subset, yielding Cohen's $\kappa = 0.82$, indicating substantial agreement.

\textbf{L3 annotation statistics.} Across all 195 samples, the mean reference minimal set size is 3.74 units (median: 3.0, range: 1--10, $\sigma$=2.54). The distribution of minimal set sizes is shown in Table~\ref{tab:l3_distribution}.

\begin{table}[htbp]
\centering
\caption{Distribution of human-annotated minimal sufficient evidence set sizes.}
\label{tab:l3_distribution}
\begin{tabular}{cccc}
\toprule
Set Size & Count & Proportion & Cumulative \\
\midrule
1 & 32 & 16.4\% & 16.4\% \\
2 & 47 & 24.1\% & 40.5\% \\
3 & 31 & 15.9\% & 56.4\% \\
4 & 31 & 15.9\% & 72.3\% \\
5 & 18 & 9.2\% & 81.5\% \\
6 & 9 & 4.6\% & 86.2\% \\
7 & 6 & 3.1\% & 89.2\% \\
8 & 4 & 2.1\% & 91.3\% \\
9 & 2 & 1.0\% & 92.3\% \\
10 & 15 & 7.7\% & 100\% \\
\bottomrule
\end{tabular}
\end{table}

Over 72\% of samples require 4 or fewer evidence units, confirming that minimal sufficient evidence sets are typically small relative to the full evidence pool (mean 28.3 units).

\textbf{Minimal set size by verdict class.} Supported claims require the fewest evidence units on average (3.25), followed by refuted (3.69) and insufficient (4.58). The larger sets for insufficient cases reflect the need for annotators to demonstrate that no compact subset suffices, often requiring evidence from multiple modalities to establish that the claim cannot be resolved.

\section{Complete Experimental Results}
\label{app:full_results}

\subsection{Per-Class F1 Scores}

Table~\ref{tab:per_class_f1} presents the per-class precision, recall, and F1 for all strategies, complementing the Macro-F1 reported in Table~\ref{tab:main_results}.

\begin{table}[htbp]
\centering
\caption{Per-class precision, recall, and F1 across all strategies (195 samples). Bold marks the best value in each column.}
\label{tab:per_class_f1}
\begin{tabular}{l ccc ccc ccc}
\toprule
 & \multicolumn{3}{c}{Supported} & \multicolumn{3}{c}{Refuted} & \multicolumn{3}{c}{Insufficient} \\
\cmidrule(lr){2-4} \cmidrule(lr){5-7} \cmidrule(lr){8-10}
Method & P & R & F1 & P & R & F1 & P & R & F1 \\
\midrule
B1 Full     & 0.632 & 0.742 & 0.682 & 0.471 & 0.615 & 0.533 & 0.500 & 0.254 & 0.337 \\
B2 Top-3    & \textbf{0.679} & 0.546 & 0.606 & 0.371 & 0.333 & 0.351 & 0.378 & \textbf{0.525} & \textbf{0.440} \\
B2 Top-5    & 0.606 & 0.649 & 0.627 & 0.405 & 0.385 & 0.395 & 0.370 & 0.339 & 0.354 \\
B2 Top-7    & 0.615 & 0.691 & 0.650 & 0.413 & 0.487 & 0.447 & 0.350 & 0.237 & 0.283 \\
B3 CoT      & 0.596 & \textbf{0.866} & \textbf{0.706} & 0.490 & \textbf{0.641} & \textbf{0.556} & \textbf{1.000} & 0.051 & 0.097 \\
B4 Greedy   & 0.605 & 0.711 & 0.654 & 0.467 & 0.538 & 0.500 & 0.333 & 0.203 & 0.253 \\
B5 Abstention & 0.641 & 0.680 & 0.660 & \textbf{0.548} & 0.436 & 0.486 & 0.377 & 0.390 & 0.383 \\
\bottomrule
\end{tabular}
\end{table}

\textbf{Key observations.}
(1) B3 achieves the highest precision on insufficient cases (1.000) because it predicts ``insufficient'' for only 3 samples---all of which happen to be correct---but its recall of 0.051 renders this precision meaningless in practice. Meanwhile, B3 achieves the highest supported recall (0.866) by aggressively committing to verdicts.
(2) B5 has the second-highest insufficient-class F1 (0.383, after B2 Top-3) while staying within 0.05 of the best supported F1 and within 0.07 of the best refuted F1. The methods with higher supported and refuted F1 (B1 and B3) are much weaker on the insufficient class (0.337 and 0.097).
(3) The refuted class is the most challenging across all methods, with the best F1 being 0.556 (B3), reflecting the difficulty of locating specific contradictory evidence. Refuted precision is consistently low across methods (0.37--0.55), indicating that when systems predict ``refuted,'' they are frequently wrong.

\subsection{Cross-Model Generalization}
\label{app:crossmodel}

To verify that our findings are not idiosyncratic to a single backbone, we re-ran the \emph{entire verification pipeline} (relevance scoring $\rightarrow$ sufficiency search $\rightarrow$ two-layer verdict) with two additional verifier models on the \emph{identical, frozen evidence pool}: the same 5{,}510 cached units (Claude-written V descriptions, Whisper ASR, retrieved E snippets), with the same unit IDs and contents. Only the verifier model changes; the inputs do not. We test \textbf{GPT-5.5} (a different provider/architecture, closed-weight) and \textbf{Qwen2.5-72B-Instruct} (open-weight). Because the descriptions remain Claude-generated, the verifier $\neq$ the description model, which also tests whether the results depend on the same model both describing and verifying. Table~\ref{tab:crossmodel_full} reports the full B1--B5 results for all three backbones.

\begin{table}[htbp]
\centering
\caption{Cross-model generalization: full results for three verifier backbones on the identical frozen evidence pool. Claude Sonnet~4 ($n{=}195$) and GPT-5.5 ($n{=}195$) cover all samples; for Qwen2.5-72B, a few samples whose claims concern disaster and violent-incident reporting were refused by the DashScope provider's content-safety filter and excluded ($n{=}193$ for B1/B2; $n{=}192$ for B4/B5). These are provider-side content refusals unrelated to the selection method. ``Units'' = average selected evidence units; ``Ins-F1'' = insufficient-class F1. Bold marks the best value per column within each backbone. The efficiency effect (B5 recovers most of B1's Macro-F1 with a fraction of the evidence) holds on GPT-5.5 and only partially on Qwen2.5-72B; the abstention gain (B5 $>$ B4 on Ins-F1) is clear on Claude Sonnet~4 and GPT-5.5 but small on Qwen2.5-72B.}
\label{tab:crossmodel_full}
\begin{tabular}{l l cccc}
\toprule
Backbone & Method & Macro-F1 & Accuracy & Units & Ins-F1 \\
\midrule
Claude Sonnet 4 & B1 Full        & \textbf{0.518} & \textbf{0.569} & 27.7 & 0.337 \\
                & B2 Top-5       & 0.459 & 0.503 & 5.0  & 0.354 \\
                & B4 Greedy      & 0.469 & 0.523 & 4.1  & 0.253 \\
                & B5 Abstention  & 0.510 & 0.544 & 4.5 & \textbf{0.383} \\
\midrule
GPT-5.5 (closed) & B1 Full        & \textbf{0.586} & \textbf{0.651} & 27.7 & 0.464 \\
                 & B2 Top-5       & 0.551 & 0.605 & 5.0  & 0.404 \\
                 & B4 Greedy      & 0.547 & 0.615 & 4.7  & 0.407 \\
                 & B5 Abstention  & 0.559 & 0.626 & 4.9 & \textbf{0.508} \\
\midrule
Qwen2.5-72B (open) & B1 Full        & \textbf{0.580} & \textbf{0.658} & 27.9 & 0.386 \\
                   & B2 Top-5       & 0.577 & 0.622 & 5.0  & \textbf{0.459} \\
                   & B4 Greedy      & 0.526 & 0.573 & 2.3  & 0.413 \\
                   & B5 Abstention  & 0.489 & 0.573 & 2.4 & 0.423 \\
\bottomrule
\end{tabular}
\end{table}

\textbf{Key observations.}
(1) \textbf{Full-evidence performance is consistent across backbones} (B1 Macro-F1 $0.52$--$0.59$); the two alternate verifiers are if anything slightly stronger than Claude, confirming that the absolute level tracks verifier capability rather than a quirk of one model.
(2) \textbf{The minimal-sufficient-subset effect replicates on GPT-5.5 and only partially on Qwen}: B5 recovers $95.3\%$ (GPT-5.5) and $84.3\%$ (Qwen) of full-evidence Macro-F1 while using only $9$--$18\%$ of the evidence units, mirroring Claude's $98.5\%$. The gap to B1 is not significant on GPT-5.5 but is significant on Qwen (Section~\ref{app:significance}).
(3) \textbf{The abstention gain scales with verifier capability.} On the two stronger verifiers (Claude, GPT-5.5), B5 clearly improves the insufficient class over B4 (Ins-F1 $+0.131$ and $+0.100$; GPT-5.5 abstention recall $0.373{\rightarrow}0.559$). The open-weight Qwen is a coarser sufficiency judge---it ends with a single evidence unit on $60\%$ of samples (vs.\ $28\%$ for GPT-5.5)---so its selected sets are smaller but its abstention margin is correspondingly thinner. The selection framework thus transfers across providers, while its benefit depends on the quality of the verifier's sufficiency judgments.

\subsection{Statistical Significance and Confidence-Threshold Baselines}
\label{app:significance}

All comparisons in this section use a paired bootstrap over videos (2{,}000 resamples, seed 42) of the difference \emph{B5 minus the compared method}, for Macro-F1 and for insufficient-class F1; positive values favor B5. In addition to the strategies of Section~\ref{sec:evidence_selection}, we evaluate \emph{confidence-threshold abstention} (``+~$\tau$''): any committed verdict (supported or refuted) whose self-reported confidence is below $\tau$ is converted to \textit{insufficient}. The main variant uses $\tau = 0.8$, the same threshold B4 and B5 use for the sufficiency check, and involves no tuning. As a tuned reference, we also select $\tau$ from $\{0, 0.05, \ldots, 0.95\}$ by 5-fold cross-validation (maximizing Macro-F1 on the training folds and applying it to the held-out fold), repeated over 20 random splits. Table~\ref{tab:significance} reports the results for all three backbones.

\begin{table}[htbp]
\centering
\small
\caption{Paired-bootstrap comparisons of B5 against every other method, on all three verifier backbones. $\Delta$ = B5 minus method (positive favors B5), with 95\% confidence intervals; $^{*}$ marks $p < 0.05$. ``+~$\tau$'' rows apply confidence-threshold abstention with $\tau = 0.8$; the last column gives the Macro-F1 of the same base method with a cross-validated $\tau$ (mean $\pm$ s.d.\ over 20 random 5-fold splits). For Qwen2.5-72B, the comparison uses the 192 samples on which all four strategies returned a verdict, so B1 and Top-5 differ slightly from Table~\ref{tab:crossmodel_full} ($n{=}193$).}
\label{tab:significance}
\setlength{\tabcolsep}{4pt}
\begin{tabular}{lccccc}
\toprule
Method & Macro-F1 & Units & $\Delta$Macro-F1 [95\% CI] & $\Delta$Ins-F1 [95\% CI] & CV-$\tau$ Macro-F1 \\
\midrule
\multicolumn{6}{l}{\textit{Claude Sonnet 4 ($n{=}195$)}} \\
B1 Full & 0.518 & 27.7 & $-$0.008 [$-$0.08, +0.06] & +0.046 [$-$0.08, +0.18] & --- \\
B2 Top-3 & 0.466 & 3.0 & +0.044 [$-$0.03, +0.12] & $-$0.056 [$-$0.18, +0.07] & --- \\
B2 Top-5 & 0.459 & 5.0 & +0.051 [$-$0.02, +0.13] & +0.029 [$-$0.11, +0.17] & --- \\
B2 Top-7 & 0.460 & 7.0 & +0.050 [$-$0.02, +0.12] & +0.101 [$-$0.02, +0.22] & --- \\
B3 CoT & 0.453 & 9.4 & +0.057 [$-$0.02, +0.13] & +0.287 [+0.16, +0.41]$^{*}$ & --- \\
B4 Greedy & 0.469 & 4.1 & +0.041 [$-$0.02, +0.10] & +0.131 [+0.03, +0.23]$^{*}$ & --- \\
B1 Full + $\tau$ & 0.525 & 27.7 & $-$0.015 [$-$0.09, +0.05] & +0.028 [$-$0.10, +0.15] & 0.534 $\pm$ 0.014 \\
B2 Top-5 + $\tau$ & 0.464 & 5.0 & +0.045 [$-$0.03, +0.12] & +0.015 [$-$0.12, +0.15] & 0.452 $\pm$ 0.009 \\
B4 Greedy + $\tau$ & 0.476 & 4.1 & +0.034 [$-$0.03, +0.09] & +0.113 [+0.01, +0.22]$^{*}$ & 0.471 $\pm$ 0.013 \\
\textit{B5 (reference)} & 0.510 & 4.5 & --- & --- & --- \\
\midrule
\multicolumn{6}{l}{\textit{GPT-5.5 ($n{=}195$)}} \\
B1 Full & 0.586 & 27.7 & $-$0.027 [$-$0.10, +0.04] & +0.043 [$-$0.05, +0.14] & --- \\
B2 Top-5 & 0.551 & 5.0 & +0.008 [$-$0.05, +0.06] & +0.104 [+0.02, +0.20]$^{*}$ & --- \\
B4 Greedy & 0.547 & 4.7 & +0.012 [$-$0.04, +0.06] & +0.100 [+0.01, +0.19]$^{*}$ & --- \\
B1 Full + $\tau$ & 0.577 & 27.7 & $-$0.018 [$-$0.08, +0.04] & +0.043 [$-$0.06, +0.13] & 0.592 $\pm$ 0.010 \\
B2 Top-5 + $\tau$ & 0.554 & 5.0 & +0.005 [$-$0.05, +0.06] & +0.032 [$-$0.05, +0.12] & 0.525 $\pm$ 0.011 \\
B4 Greedy + $\tau$ & 0.527 & 4.7 & +0.032 [$-$0.03, +0.09] & +0.054 [$-$0.03, +0.13] & 0.564 $\pm$ 0.007 \\
\textit{B5 (reference)} & 0.559 & 4.9 & --- & --- & --- \\
\midrule
\multicolumn{6}{l}{\textit{Qwen2.5-72B ($n{=}192$)}} \\
B1 Full & 0.575 & 27.9 & $-$0.086 [$-$0.17, $-$0.00]$^{*}$ & +0.052 [$-$0.09, +0.20] & --- \\
B2 Top-5 & 0.574 & 5.0 & $-$0.085 [$-$0.16, $-$0.01]$^{*}$ & $-$0.026 [$-$0.14, +0.09] & --- \\
B4 Greedy & 0.526 & 2.3 & $-$0.037 [$-$0.09, +0.02] & +0.010 [$-$0.04, +0.06] & --- \\
B1 Full + $\tau$ & 0.590 & 27.9 & $-$0.101 [$-$0.19, $-$0.01]$^{*}$ & +0.013 [$-$0.13, +0.17] & 0.566 $\pm$ 0.020 \\
B2 Top-5 + $\tau$ & 0.574 & 5.0 & $-$0.085 [$-$0.16, $-$0.01]$^{*}$ & $-$0.022 [$-$0.13, +0.09] & 0.563 $\pm$ 0.013 \\
B4 Greedy + $\tau$ & 0.526 & 2.3 & $-$0.037 [$-$0.09, +0.02] & +0.010 [$-$0.04, +0.06] & 0.509 $\pm$ 0.007 \\
\textit{B5 (reference)} & 0.489 & 2.4 & --- & --- & --- \\
\bottomrule
\end{tabular}
\end{table}

\textbf{Key observations.}
(1) \textbf{B5 matches full evidence on the two stronger verifiers.} On Claude Sonnet~4 and GPT-5.5, no Macro-F1 difference between B5 and either B1 or thresholded B1 is significant, although B5 uses 16--18\% of the evidence. On Qwen2.5-72B, B5 is significantly below B1 and Top-5, consistent with Section~\ref{app:crossmodel}: Qwen's coarse sufficiency judgments stop the search too early (2.4 units on average).
(2) \textbf{Margins among efficient methods are not significant at $n{=}195$.} On Claude, B5's Macro-F1 exceeds every efficient strategy (Top-K, CoT, B4, and their thresholded variants) by 0.03--0.06, but none of these margins is statistically significant; on GPT-5.5 the efficient methods lie within 0.04 of each other.
(3) \textbf{Abstention gains concentrate in the insufficient class.} B5's significant advantages are in insufficient-class F1: over B3 and B4 on Claude, over Top-5 and B4 on GPT-5.5, and over thresholded B4 on Claude ($+0.113$). On Claude and GPT-5.5, thresholding lifts B1, Top-5, and B4 toward B5 on this class, which is why we regard explicit abstention, rather than any particular abstention mechanism, as the key ingredient (Section~\ref{sec:main_results}).
(4) \textbf{Tuning $\tau$ does not change the conclusions.} Median cross-validated thresholds lie between 0.75 and 0.85 for B1 and B4 on Claude and GPT-5.5, close to the fixed $\tau = 0.8$. On the first cross-validation split, B5 does not differ significantly from tuned B1 (Claude: $-0.011$, $p{=}0.78$; GPT-5.5: $-0.040$, $p{=}0.25$) or tuned B4 (Claude: $+0.021$, $p{=}0.55$; GPT-5.5: $-0.007$, $p{=}0.79$).

\subsection{Confusion Matrices}

We present confusion matrices for B5 (proposed) and B1 (full-evidence baseline).

\begin{table}[htbp]
\centering
\caption{Confusion matrix for B5 (Greedy + Abstention).}
\label{tab:cm_b5}
\begin{tabular}{l ccc}
\toprule
True $\backslash$ Pred & Supported & Refuted & Insufficient \\
\midrule
Supported (97)   & \textbf{66} & 8 & 23 \\
Refuted (39)     & 7 & \textbf{17} & 15 \\
Insufficient (59)& 30 & 6 & \textbf{23} \\
\bottomrule
\end{tabular}
\end{table}

\begin{table}[htbp]
\centering
\caption{Confusion matrix for B1 (Full Evidence).}
\label{tab:cm_b1}
\begin{tabular}{l ccc}
\toprule
True $\backslash$ Pred & Supported & Refuted & Insufficient \\
\midrule
Supported (97)   & \textbf{72} & 16 & 9 \\
Refuted (39)     & 9 & \textbf{24} & 6 \\
Insufficient (59)& 33 & 11 & \textbf{15} \\
\bottomrule
\end{tabular}
\end{table}

\textbf{Analysis.} B1 correctly identifies more supported cases (72 vs.\ 66) and more refuted cases (24 vs.\ 17) due to its access to all evidence, but performs worse on insufficient cases (15 vs.\ 23). The key difference is in the insufficient row: B1 misclassifies 33 insufficient samples as supported (overconfident), while B5 correctly identifies 23 insufficient cases. B5's abstention mechanism converts potential overconfident misclassifications into calibrated ``insufficient'' predictions, at the cost of more supported samples being conservatively labeled as insufficient (23 vs.\ 9).

\subsection{Consistency Layer Distribution}

Table~\ref{tab:consistency_dist} reports the distribution of Layer~1 (consistency) outputs across strategies, showing how each method assesses claim--video alignment.

\begin{table}[htbp]
\centering
\caption{Distribution of Layer~1 (consistency) assessments across strategies.}
\label{tab:consistency_dist}
\begin{tabular}{lccc}
\toprule
Method & Consistent & Inconsistent & Unclear \\
\midrule
B1 Full     & 168 (86.2\%) & 18 (9.2\%) & 9 (4.6\%) \\
B2 Top-5    & 117 (60.0\%) & 9 (4.6\%)  & 69 (35.4\%) \\
B3 CoT      & 141 (72.3\%) & 54 (27.7\%) & 0 (0.0\%) \\
B4 Greedy   & 83 (42.6\%) & 13 (6.7\%) & 99 (50.8\%)$^\dagger$ \\
B5 Abstention & 79 (40.5\%) & 14 (7.2\%) & 102 (52.3\%)$^\dagger$ \\
\bottomrule
\end{tabular}

\smallskip
\begin{minipage}{0.9\linewidth}\footnotesize $^\dagger$Most ``unclear'' labels for B4 and B5 arise because the pruned subset contains only external units, leaving Layer~1 no internal evidence to assess (68 of 99 for B4; 67 of 102 for B5). For the 39 samples on which B5 abstains, the label comes from the last sufficiency check, because these samples skip the two-layer verification.
\end{minipage}
\end{table}

B3 produces zero ``unclear'' assessments and the most ``inconsistent'' assessments (27.7\%), reflecting its tendency to overcommit to strong judgments in both layers. B2 Top-5 produces a high ``unclear'' rate (35.4\%) because fixed-budget selection may exclude internal evidence in favor of higher-scoring external units, depriving Layer~1 of the visual and speech signals needed for consistency assessment.

\subsection{Ablation: Per-Class Results}

Table~\ref{tab:ablation_perclass} expands the ablation study with per-class F1 scores. As in Section~\ref{sec:ablation}, the abstention gate is ablated by comparing B5 with B4, and all other components are ablated on B4; the reference for those rows is therefore B4.

\begin{table}[htbp]
\centering
\caption{Per-class F1 for ablation variants. Rows below the rule are ablations of B4 (the selection pipeline without the abstention gate) and should be compared with the B4 row.}
\label{tab:ablation_perclass}
\begin{tabular}{lcccr}
\toprule
Setting & F1-Sup & F1-Ref & F1-Ins & Macro-F1 \\
\midrule
B5 (proposed) & 0.660 & 0.486 & 0.383 & 0.510 \\
B4 (= B5 w/o Abstention) & 0.654 & 0.500 & 0.253 & 0.469 \\
\midrule
B4 w/o External & 0.096 & 0.270 & 0.453 & 0.273 \\
B4 w/o V+A (E-only) & 0.660 & 0.453 & 0.340 & 0.484 \\
B4 w/o ASR & 0.638 & 0.494 & 0.263 & 0.465 \\
B4 w/o Pruning & 0.660 & 0.506 & 0.265 & 0.477 \\
\bottomrule
\end{tabular}
\end{table}

\textbf{Key observations.}
(1) The abstention gate acts almost entirely on the insufficient class: compared with B4, B5 raises insufficient F1 from 0.253 to 0.383 while supported and refuted F1 change by less than 0.015.
(2) Removing external evidence causes the most dramatic degradation: relative to B4, supported F1 collapses from 0.654 to 0.096 and refuted F1 from 0.500 to 0.270, while insufficient F1 rises to 0.453 because the verifier returns \textit{insufficient} for most samples even without an abstention gate---consistent with the finding in Section~\ref{sec:ablation} that external evidence is indispensable for factual determination.
(3) Without internal evidence (E-only), Macro-F1 stays close to B4 (0.484 vs.\ 0.469), with per-class shifts in both directions (supported 0.660 vs.\ 0.654, refuted 0.453 vs.\ 0.500, insufficient 0.340 vs.\ 0.253), but 100\% of samples receive ``unclear'' in Layer~1, confirming that V+A are necessary for interpretable consistency assessment. Supported F1 is unchanged even without internal evidence, indicating that external sources carry the dominant signal for verifying supported claims.
(4) Removing ASR alone or skipping pruning changes no class by more than 0.02 relative to B4. Pruning therefore makes the evidence set more compact (4.1 vs.\ 5.4 units on average) without measurable cost to any verdict class.

\section{Case Study}
\label{app:case_study}

We present two cases illustrating B5's strengths and limitations.

\subsection{Success Case: manual\_0034}

\begin{itemize}
\item \textbf{Claim:} A Sydney tech entrepreneur's adopted dog, after failed conventional treatment, had its tumor halved by an anti-cancer drug created using ChatGPT and AlphaFold.
\item \textbf{Ground truth:} Supported
\item \textbf{Total evidence units:} 74 (metadata excluded)
\end{itemize}

\textbf{B5 selects 4 units out of 74 (5.4\%) and correctly outputs ``supported'' (conf.\ 0.85).}

The selected evidence set:
\begin{itemize}
\item \textbf{A2} (ASR): Garbled transcription mentioning ``ChatGPT and AlphaFold made a dog's anti-cancer drug''---weak but topically anchoring.
\item \textbf{E1} (External, MIT Technology Review China): Confirms Paul used ChatGPT to brainstorm treatments and designed an mRNA cancer vaccine for his dog.
\item \textbf{E4} (External, yzwb.net): Independent confirmation from Paul Conyngham; tumor shrank significantly.
\item \textbf{E5} (External, Zhihu): Confirms ChatGPT + AlphaFold pipeline; DNA sequencing cost only \$3,000.
\end{itemize}

\textbf{Human-annotated minimal set:} \{A2, E1, E4, E5\}---exactly the same 4 units, yielding \textbf{100\% overlap} (Jaccard = 1.0).

\textbf{Comparison across strategies:}
\begin{itemize}
\item B1 (Full): correct, used all 74 units (100\%)
\item B3 (CoT): correct, self-selected 21 units (28.4\%)
\item B4 (Greedy): correct, used 4 units (5.4\%)
\item B5 (Abstention): correct, used 4 units (5.4\%)
\end{itemize}

This case illustrates one of B5's strengths: the selected evidence coincides with the set independently identified by human annotators as minimally sufficient, achieving a 94.6\% evidence reduction without loss of verdict accuracy. The selected set combines one weak internal signal (garbled ASR) with three independent external confirmations, consistent with the two-layer framework's design where external evidence drives factual determination.

\subsection{Failure Case: manual\_0059}

\begin{itemize}
\item \textbf{Claim:} US company ``Newkins'' launches the world's first custom baby service---\$43K to screen embryos for IQ, appearance, and lifespan across 900+ traits; tested on 120K embryos; used by Musk and Thiel.
\item \textbf{Ground truth:} Refuted
\item \textbf{Total evidence units:} 32
\end{itemize}

\textbf{B5 selects 10 units (the full candidate pool) but outputs ``insufficient'' instead of ``refuted'' (conf.\ 0.3).}

The human-annotated minimal set contains 5 units \{V2, V7, A1, A3, E5\}. B5's selected set of 10 units is a \textbf{superset} of the human minimal set (5/5 covered), yet the system fails to synthesize them into a refutation.

B1 (Full Evidence) correctly identifies ``refuted'' (conf.\ 0.85) by cross-referencing multiple evidence pieces: (1) the real company is ``Nucleus Genomics,'' not ``Newkins''; (2) it screens only $\sim$12 diseases, not 900+ traits; (3) actual cost is \$2,500 per embryo, not \$43K total; (4) no evidence that Musk or Thiel used the service; (5) the technology is a commercialization of existing PGT-P, not ``the world's first.''

\textbf{Comparison across strategies:}
\begin{itemize}
\item B1 (Full): \textbf{correct} (refuted), used all 32 units
\item B3 (CoT): \textbf{correct} (refuted), self-selected evidence
\item B4 (Greedy): incorrect (insufficient), used 10 units
\item B5 (Abstention): incorrect (insufficient), used 10 units
\end{itemize}

This case suggests that B5's failure may stem less from evidence \emph{coverage} than from the difficulty of synthesizing multiple partially contradictory pieces into a confident refutation. The claim mixes real elements (a real company, real technology) with fabrications and exaggerations, requiring cross-referencing across multiple evidence pieces to identify the pattern of distortion. B5's incremental greedy process evaluates sufficiency after each addition, but refutation of such complex claims may require holistic reasoning over the full evidence constellation---something the full-evidence baseline B1 achieves by processing all evidence jointly.

\section{Dataset Statistics}
\label{app:dataset_stats}

\subsection{Topic Distribution}

Table~\ref{tab:topic_dist} shows the distribution of videos across topic categories, annotated via keyword-based classification.

\begin{table}[htbp]
\centering
\caption{Topic distribution of MiniVer-V samples.}
\label{tab:topic_dist}
\begin{tabular}{lcc}
\toprule
Topic & Count & Proportion \\
\midrule
International Politics \& Military & 45 & 23.1\% \\
Domestic News \& Society & 30 & 15.4\% \\
Health \& Medicine & 29 & 14.9\% \\
Science \& Technology & 24 & 12.3\% \\
Natural Disaster \& Accident & 22 & 11.3\% \\
Food Safety \& Consumer & 14 & 7.2\% \\
Sports \& Entertainment & 13 & 6.7\% \\
Daily Life \& Myth-busting & 12 & 6.2\% \\
Other & 6 & 3.1\% \\
\midrule
\textbf{Total} & \textbf{195} & \textbf{100\%} \\
\bottomrule
\end{tabular}
\end{table}

The dataset spans a broad range of topics, with international politics being the most represented (23.1\%), reflecting the prevalence of geopolitical misinformation on Chinese social media platforms during the data collection period. Health \& Medicine (14.9\%) and Science \& Technology (12.3\%) are also well-represented, consistent with prior findings that these domains are frequent targets of online misinformation.

\subsection{Platform Distribution}
\label{app:platform}

Table~\ref{tab:platform_dist} shows the source platform of each video. Most videos come from the two largest Chinese short-video platforms, Douyin and Xiaohongshu; the remainder come from X, Bilibili, Weibo, and the video channels of news outlets. All claims are written in Chinese; 40 videos (20.5\%) contain mainly English speech, defined as CJK characters making up at most 10\% of the ASR transcript.

\begin{table}[htbp]
\centering
\caption{Source platform distribution of MiniVer-V samples.}
\label{tab:platform_dist}
\begin{tabular}{lcc}
\toprule
Platform & Count & Proportion \\
\midrule
Douyin & 112 & 57.4\% \\
Xiaohongshu & 64 & 32.8\% \\
X (Twitter) & 8 & 4.1\% \\
Bilibili & 3 & 1.5\% \\
Weibo & 3 & 1.5\% \\
News outlets (CNN, BBC, Xinhua, Southern Daily) & 5 & 2.6\% \\
\midrule
\textbf{Total} & \textbf{195} & \textbf{100\%} \\
\bottomrule
\end{tabular}
\end{table}

\subsection{Evidence Unit Distribution}

Table~\ref{tab:evidence_modality_stats} presents detailed per-modality statistics after metadata exclusion.

\begin{table}[htbp]
\centering
\caption{Evidence unit statistics by modality (metadata excluded).}
\label{tab:evidence_modality_stats}
\begin{tabular}{lcccc}
\toprule
Modality & Total & Mean/Sample & Proportion & Source \\
\midrule
Visual (V) & 1,207 & 6.2 & 21.9\% & Claude Sonnet 4 \\
ASR (A) & 2,847 & 14.6 & 51.7\% & Whisper (base) \\
External (E) & 1,456 & 7.5 & 26.4\% & Web Search \\
\midrule
\textbf{Total} & \textbf{5,510} & \textbf{28.3} & \textbf{100\%} & \\
\bottomrule
\end{tabular}
\end{table}

\subsection{Modality Composition in Human-Annotated Minimal Sets}

Table~\ref{tab:minimal_modality} shows the modality composition of the human-annotated minimal sufficient evidence sets (L3 annotation), providing insight into which evidence types humans consider most critical for verification.

\begin{table}[htbp]
\centering
\caption{Modality composition in human-annotated minimal sufficient sets.}
\label{tab:minimal_modality}
\begin{tabular}{lccc}
\toprule
Modality & Units Selected & Proportion & Contrast with Pool \\
\midrule
Visual (V) & 116 & 15.9\% & 21.9\% in pool \\
ASR (A) & 237 & 32.5\% & 51.7\% in pool \\
External (E) & 376 & 51.6\% & 26.4\% in pool \\
\midrule
\textbf{Total} & \textbf{729} & \textbf{100\%} & \\
\bottomrule
\end{tabular}
\end{table}

External evidence is overrepresented in minimal sets (51.6\% selected vs.\ 26.4\% in pool), consistent with its critical role in factual determination as shown by the ablation results in Section~\ref{sec:ablation}. Conversely, ASR is underrepresented (32.5\% vs.\ 51.7\%), suggesting that while speech content is voluminous, much of it is redundant for verification purposes.

\subsection{Evidence Efficiency by Strategy}

Table~\ref{tab:efficiency_detail} provides a detailed breakdown of evidence usage across correctly and incorrectly classified samples.

\begin{table}[htbp]
\centering
\caption{Evidence efficiency: average units used for correct vs.\ incorrect predictions.}
\label{tab:efficiency_detail}
\begin{tabular}{lccc}
\toprule
Method & Avg (All) & Avg (Correct) & Avg (Incorrect) \\
\midrule
B1 Full & 27.7 & 27.7 & 27.7 \\
B2 Top-3 & 3.0 & 3.0 & 3.0 \\
B2 Top-5 & 5.0 & 5.0 & 5.0 \\
B2 Top-7 & 7.0 & 7.0 & 7.0 \\
B3 CoT & 9.4 & 9.2 & 9.6 \\
B4 Greedy & 4.1 & 3.5 & 4.8 \\
B5 Abstention & 4.5 & 3.8 & 5.2 \\
\bottomrule
\end{tabular}
\end{table}

For B4 and B5, correctly classified samples use substantially fewer evidence units on average (3.5/3.8) than incorrectly classified ones (4.8/5.2). This gap suggests that ``easy'' samples reach sufficiency quickly with a small evidence set, while ``hard'' samples accumulate more evidence without reaching the sufficiency threshold---the model exhausts a larger portion of the candidate pool on samples it ultimately gets wrong. For B1 and B2, evidence counts are fixed by design and thus identical across correct and incorrect predictions.

\section{Additional Metric Definitions}
\label{app:metrics}

For completeness, we provide formal definitions for the metrics used or discussed in our analysis, including several evidence-selection diagnostics not reported in the main text.

\subsection{Verdict Macro-F1}

The primary evaluation metric. Let $\mathcal{C} = \{\text{supported}, \text{refuted}, \text{insufficient}\}$ be the set of verdict classes. For each class $c \in \mathcal{C}$, we compute precision $P_c$, recall $R_c$, and F1 score $F_{1,c}$:
\begin{align}
P_c &= \frac{TP_c}{TP_c + FP_c} \\
R_c &= \frac{TP_c}{TP_c + FN_c} \\
F_{1,c} &= \frac{2 \cdot P_c \cdot R_c}{P_c + R_c}
\end{align}
where $TP_c$, $FP_c$, and $FN_c$ are the true positives, false positives, and false negatives for class $c$, respectively. The Macro-F1 is the unweighted average:
\begin{equation}
\text{Macro-F1} = \frac{1}{|\mathcal{C}|} \sum_{c \in \mathcal{C}} F_{1,c}
\end{equation}

Macro-F1 treats all classes equally regardless of sample count, penalizing methods that sacrifice minority-class performance (e.g., refuted or insufficient) for majority-class gains.

\subsection{Consistency Accuracy}

Measures the accuracy of the Layer~1 consistency assessment against human annotations:
\begin{equation}
\text{ConsAcc} = \frac{|\{i : \hat{y}^{\text{cons}}_i = y^{\text{cons}}_i\}|}{N}
\end{equation}
where $\hat{y}^{\text{cons}}_i$ and $y^{\text{cons}}_i$ are the predicted and gold consistency labels for sample $i$, and $N$ is the number of samples with gold consistency annotations. This metric would require gold consistency annotations and is included here for completeness; it is not reported in the current study.

\subsection{Sufficiency Accuracy}

Measures whether the model's selected evidence covers the human-annotated minimal sufficient set. For samples with gold verdict $\neq$ insufficient:
\begin{equation}
\text{SufAcc} = \frac{|\{i : \mathcal{S}^*_i \subseteq \hat{\mathcal{S}}_i\}|}{N_{\text{non-insuf}}}
\end{equation}
where $\mathcal{S}^*_i$ is the human-annotated minimal set and $\hat{\mathcal{S}}_i$ is the model's selected evidence. Sufficiency accuracy checks that the model's selection is a superset of the human reference.

\subsection{Minimality Rate}

Evaluates how compact the model's selections are relative to the human reference:
\begin{equation}
\text{MinPrec}_i = \frac{|\hat{\mathcal{S}}_i \cap \mathcal{S}^*_i|}{|\hat{\mathcal{S}}_i|}, \quad
\text{SizeRatio}_i = \frac{|\hat{\mathcal{S}}_i|}{|\mathcal{S}^*_i|}
\end{equation}

Minimality Precision measures the proportion of model-selected units that appear in the human reference (higher is better). Size Ratio indicates how many times larger the model's selection is compared to the human reference (closer to 1.0 is better; $>1$ indicates redundancy).

\subsection{Evidence Efficiency}

The average number of evidence units used per sample:
\begin{equation}
\text{AvgUnits} = \frac{1}{N} \sum_{i=1}^{N} |\hat{\mathcal{S}}_i|
\end{equation}

We also report this metric separately for correctly and incorrectly classified samples to analyze whether evidence quantity correlates with prediction quality.

\subsection{Abstention Quality}

Evaluated through three sub-metrics focused on the ``insufficient'' prediction:
\begin{align}
\text{Abst. Precision} &= \frac{TP_{\text{ins}}}{TP_{\text{ins}} + FP_{\text{ins}}} \\
\text{Abst. Recall} &= \frac{TP_{\text{ins}}}{TP_{\text{ins}} + FN_{\text{ins}}} \\
\text{OverConf} &= 1 - \text{Abst. Recall} = \frac{FN_{\text{ins}}}{TP_{\text{ins}} + FN_{\text{ins}}}
\end{align}
where $TP_{\text{ins}}$ counts samples correctly predicted as insufficient, $FP_{\text{ins}}$ counts non-insufficient samples incorrectly predicted as insufficient, and $FN_{\text{ins}}$ counts truly insufficient samples predicted as non-insufficient.

The Overconfidence Rate specifically measures the system's tendency to fail to abstain when evidence is genuinely inadequate---a particularly dangerous failure mode in high-stakes fact-checking applications.

\subsection{Evidence Faithfulness}

Measures whether the model's verdict is grounded in evidence that overlaps with the human-annotated reference:
\begin{equation}
\text{Faithfulness} = \frac{|\{i : \hat{\mathcal{S}}_i \cap \mathcal{S}^*_i \neq \emptyset, \ \hat{y}_i \neq \text{insuf.}\}|}{|\{i : \hat{y}_i \neq \text{insuf.}\}|}
\end{equation}

A verdict is considered ``faithful'' if the model's selected evidence set contains at least one unit from the human-annotated minimal set. Low faithfulness indicates that the model reaches correct verdicts through evidence not considered critical by human annotators---a potential interpretability concern.

\section{Model Configuration}
\label{app:config}

For completeness, we list the full model configuration used across all experiments.

\medskip
\noindent
\begin{center}
\begin{tabular}{lp{0.5\columnwidth}}
\toprule
Parameter & Value \\
\midrule
\multicolumn{2}{l}{\textit{Main experiments}} \\
Backbone model & Claude Sonnet 4 \\
Model ID & claude-sonnet-4-20250514 \\
Provider & Anthropic \\
Temperature & 0.1 \\
Max generation tokens & 4,096 \\
ASR model & Whisper (base) \\
Visual description model & Claude Sonnet 4 \\
Search API & SerpAPI (Google Search) \\
Candidate pool size (B4/B5) & 10 \\
Sufficiency threshold $\tau$ & 0.8 \\
Min.\ internal units in pool & 2 \\
Min.\ external units in pool & 3 \\
Max.\ external units (B1) & 8 \\
Unit truncation in prompts & 150 characters per unit (100 for relevance scoring) \\
\midrule
\multicolumn{2}{l}{\textit{Cross-model verifiers (Section~\ref{sec:crossmodel})}} \\
GPT-5.5 & OpenAI API, model \texttt{gpt-5.5}; default temperature (reasoning model); max.\ 4,096 completion tokens \\
Qwen2.5-72B-Instruct & DashScope API (OpenAI-compatible), model \texttt{qwen2.5-72b-instruct}; temperature 0.1; max.\ 4,096 tokens \\
Scoring and sufficiency checks & Same model as the verifier \\
Evidence pool & Identical frozen pool (Claude-generated V descriptions) \\
\midrule
\multicolumn{2}{l}{\textit{Raw-keyframe control (Section~\ref{sec:robustness})}} \\
Verifier & Claude Sonnet 4 (B1) \\
Visual input & Keyframe images (JPEG) in place of V descriptions \\
Subset & 60 videos, 20 per verdict class \\
\bottomrule
\end{tabular}
\end{center}

\end{document}